\documentclass[10pt,twocolumn,letterpaper]{article}

\usepackage[pagenumbers]{cvpr} 

\definecolor{cvprblue}{rgb}{0.21,0.49,0.74}
\usepackage[pagebackref,breaklinks,colorlinks,allcolors=cvprblue]{hyperref}
\usepackage[symbol]{footmisc}

\def\paperID{} 
\def\confName{CVPR}
\def\confYear{2026}

\newcommand{\ours}{Re$^2$MoGen}

\title{Re$^2$MoGen: Open-Vocabulary Motion Generation via LLM Reasoning and Physics-Aware Refinement}

\author{
Jiakun Zheng$^1$, Ting Xiao$^1$\footnotemark[1], Shiqin Cao$^1$, Xinran Li$^2$, Zhe Wang$^1$, 
Chenjia Bai$^3$\footnotemark[1]\\ \\
$^1$East China University of Science and Technology\\
$^2$Hong Kong University of Science and Technology\\
$^3$Institute of Artificial Intelligence (TeleAI), China Telecom\\
{\tt\small \{zhengjk, caoshiqin\}@mail.ecust.edu.cn, \{xiaoting, wangzhe\}@ecust.edu.cn,}\\
{\tt\small xinran.li@connect.ust.hk, bai\_chenjia@163.com}
}

\begin{document}
\maketitle

\footnotetext{* Corresponding author} 
\begin{abstract}
Text-to-motion (T2M) generation aims to control the behavior of a target character via textual descriptions. Leveraging text-motion paired datasets, existing T2M models have achieved impressive performance in generating high-quality motions within the distribution of their training data. However, their performance deteriorates notably when the motion descriptions differ significantly from the training texts. To address this issue, we propose \ours, a \textbf{Re}asoning and \textbf{Re}finement open-vocabulary \textbf{Mo}tion \textbf{Gen}eration framework that leverages enhanced Large Language Model (LLM) reasoning to generate an initial motion planning and then refine its physical plausibility via reinforcement learning (RL) post-training. Specifically, \ours{} consists of three stages:  We first employ Monte Carlo tree search to enhance the LLM's reasoning ability in generating reasonable keyframes of the motion based on text prompts, specifying only the root and several key joints’ positions to ease the reasoning process. Then, we apply a human pose model as a prior to optimize the full-body poses based on the planned keyframes and use the resulting incomplete motion to supervise fine-tuning a pre-trained motion generator via a dynamic temporal matching objective, enabling spatiotemporal completion. Finally, we use post-training with physics-aware reward to refine motion quality to eliminate physical implausibility in LLM-planned motions. 
Extensive experiments demonstrate that our framework can generate semantically consistent and physically plausible motions and achieve state-of-the-art performance in open-vocabulary motion generation. 
\end{abstract}
\section{Introduction}
\label{sec:intro}

Text-driven human motion generation has shown great promise in controlling animated characters~\cite{tevet2025closd, Zhao2025DART, PHC} and humanoid robots~\cite{UH1, li2026pchc, wang2026halo, liang2026interreal, wang2026x, lin2026pro, xie2026textop}. While most existing methods~\cite{mdm, mld, guo2024momask, jiang2024motiongpt} excel at generating high-quality, semantically consistent motions within the confines of their training data, they often fail when encountering with open-vocabulary descriptions beyond the training distribution. This challenge arises from the inherent disparity between finite motion data and infinite linguistic diversity, making it difficult to ensure both fidelity to novel descriptions and motion quality.

To address the open-vocabulary motion generation challenge, a line of work~\cite{hong2022avatarclip, tevet2022motionclip} maps motion data into the visual-language embedding space, e.g., CLIP~\cite{radford2021clip}, leveraging its cross-modal alignment capability for open-vocabulary generation. Nevertheless, such approaches typically deconstruct motion sequences into individual static poses for alignment, discarding crucial temporal dynamics and thereby frequently generating unrealistic or disjointed motions. Another line of work~\cite{pro-moiton, mandelli2025generation} employs LLMs to decompose unseen action descriptions into combinations of seen textual elements, then utilizes pre-trained models for generation and splicing. However, these methods remain constrained by the textual coverage of pre-training data, and their staged planning and splicing process is prone to error accumulation, causing the generated actions to deviate from the initial text description. Additionally, some research~\cite {cui2024anyskill,cui2025grove} exploits the exploration capability of agents in reinforcement learning (RL), guiding them via carefully designed reward functions to discover action states conforming to the text. Yet, these online exploration approaches tend to yield unstable results and critically depend on the quality of the reward design.

Given these fundamental limitations in existing approaches, a novel open-vocabulary generation method is worth discussing. Recent breakthroughs in LLMs have demonstrated remarkable progress in spatial reasoning and action planning capabilities~\cite{ji2025robobrain,azzolini2025cosmos}, enabling their successful application to diverse planning tasks~\cite{jing2025humanoidgen}. Capitalizing on these advancements, we propose \ours, a novel open-vocabulary motion generation framework based on LLM reasoning and physics-aware refinement. \ours~consists of three stages. First, we employ Monte Carlo Tree Search (MCTS)-enhanced LLM reasoning to plan a keyframe sequence that captures the core actions of the given motion description. To simplify such a planning task, the sequence only contains positions of the root joint (pelvis) and four end-effector joints (left/right wrist \& left/right ankle). Subsequently, a human pose model is used as a prior to optimize full-body poses for the LLM-planned keyframes. The resulting incomplete motion is then used to supervise fine-tuning on a pre-trained motion generation model via a dynamic temporal matching objective based on soft Dynamic Time Warping (soft-DTW), leveraging motion prior knowledge of the pre-trained model to complete the key poses into a full motion sequence spatiotemporally. Finally,~to address physical implausibility issues (e.g., foot sliding, ground penetration, and floating artifacts) in LLM-planned motions, we adopt RL post-training with physics-aware rewards to refine the generated motions.

To evaluate the effectiveness of \ours, we reconstruct a dataset based on HumanML3D \cite{CVPR2022hml3d} to train the motion generation model. Then, we collect a set of unseen text descriptions based on the semantic similarity of the texts in the dataset. 
We conduct the above motion generation process on these unseen texts to evaluate both the text-to-motion alignment and physical plausibility of the generated motions. Experimental results demonstrate that our proposed framework effectively generates semantically aligned and physically plausible motion sequences for open-vocabulary descriptions, outperforming existing open-vocabulary motion generation methods. 

The main contribution of this paper is summarized as follows: (i)~We propose a novel LLM-driven motion generation algorithm to enhance the generalization ability of the motion generation models. (ii)~We combine both supervised finetuning and RL post-training to address the semantic rationality and physical plausibility of generated motions. (iii)~We evaluate the effectiveness of our method, and the results show superior performance on challenging unseen texts.  

\section{Related Work}
\label{sec:rel}

\subsection{Open-vocabulary Motion Generation} 
Recent advances in open-vocabulary motion generation aim to bridge the gap between natural language descriptions and human motion generation for out-of-distribution text prompts. While traditional motion generation methods~\cite{mdm,mld,guo2024momask,jiang2024motiongpt,zhang2023generating} learn text-motion relationships within constrained datasets, they fail to generalize to unseen text descriptions. MotionCLIP~\cite{tevet2022motionclip} and AvatarCLIP~\cite{hong2022avatarclip} address this by aligning motions with CLIP~\cite{radford2021clip} through static pose rendering. However, they often neglect temporal dynamics, leading to unrealistic or discontinuous motion sequences. PRO-Motion~\cite{pro-moiton}, MCD~\cite{mandelli2025mcd}, and DSO-Net~\cite{fan2024DSO} leverage LLMs to decompose unseen motion descriptions into known textual components and generate corresponding key poses or motions via text-to-pose/motion models, thus remaining constrained by the underlying text-pose/motion datasets. Another line of RL-based approaches, such as AnySkill~\cite{cui2024anyskill} and GROVE~\cite{cui2025grove}, employ semantically-guided reward functions to steer agents toward text-aligned behaviors through exploration. However, these methods often suffer from unstable outcomes. Differently, our method eliminates the dependency on constrained datasets through MCTS-enhanced LLM reasoning, enabling effective planning of reasonable initial keyframes for open-vocabulary text descriptions. Furthermore, through full-body optimization and action completion, coupled with RL-based post-training, our model can generate spatiotemporally coherent and physically plausible motions.

\subsection{LLM-based Spatial Planning}
Recent advances in LLMs~\cite{brown2020language, dagan2023dynamic} have extended their capabilities beyond pure language tasks to spatial reasoning and structured environment planning. Piror researches employ~\cite{zhou2024navgpt, shah2022roboticnav} employ LLMs to parse natural language instructions into structured waypoints for navigation tasks. Other studies~\cite{brohan2023can, singh2022progprompt, song2023llmlanner} utilize LLMs' spatial understanding capabilities for embodied task decomposition and planning. More recent approaches~\cite{feng2024layoutgpt, xia2024scenegenagent} explore LLM-aided design synthesis, employing LLMs to propose spatial layouts. However, most of these LLM-based approaches focus on LLMs' abstract-level spatial understanding, such as `put the banana on the plate', rather than concrete spatial planning like `move the hand forward $0.1$m then upward $0.2$m'. In contrast, our work fully leverages LLMs' spatial planning capability to generate precise spatial coordinates from textual descriptions.
\begin{figure*}[t]
    \centering
    \includegraphics[width=0.9 \linewidth]{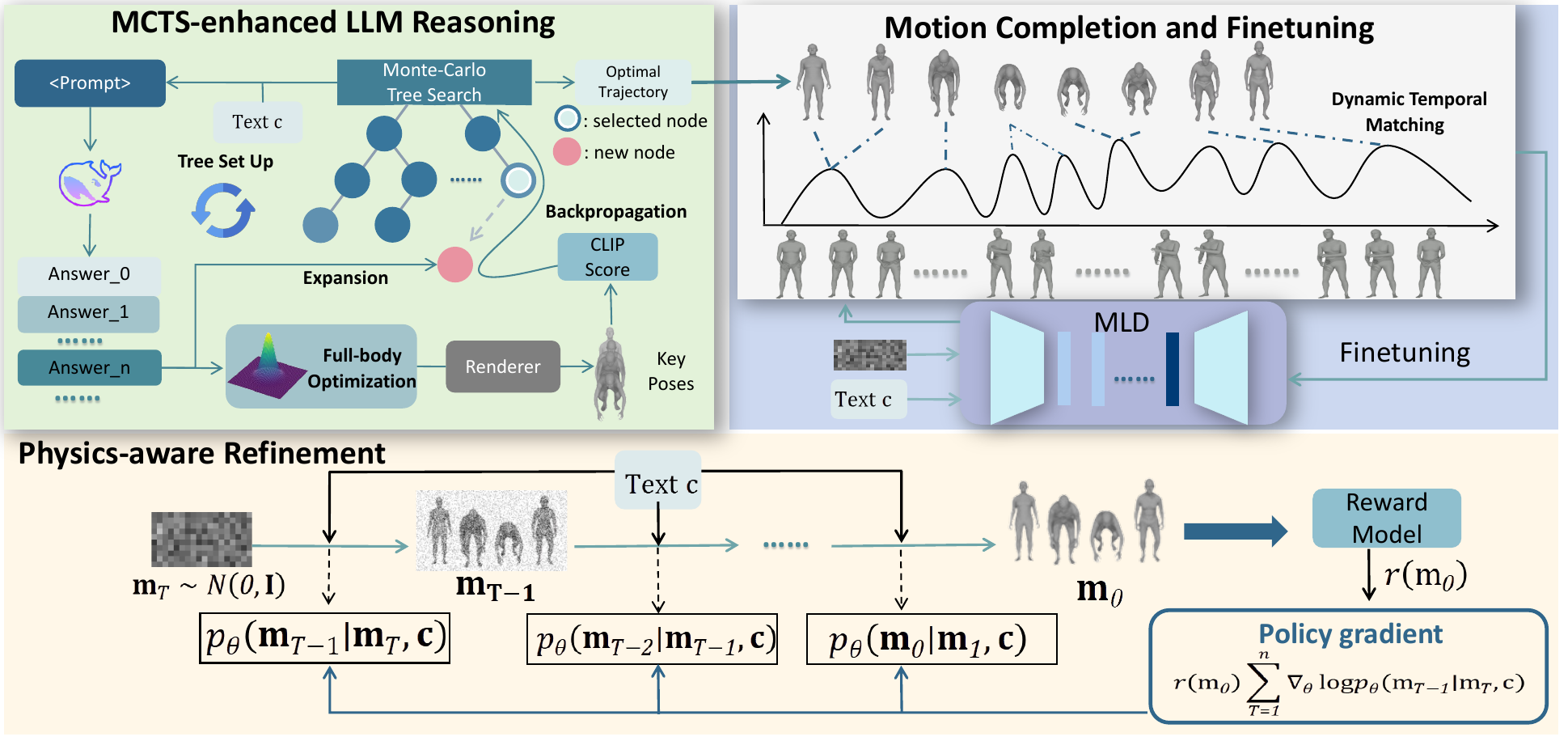}
    \caption{The framework of \ours{}, which consists of three key parts: (i) MCTS-enhanced LLM Reasoning; (ii) Motion Completion and Finetuning, and (iii) Physics-aware refinement.}
    \label{fig:overview}
    \vspace{-1em}
\end{figure*}

\section{Method}

\subsection{Preliminaries}

We give the basic process of motion generation as follows. Given a sentence of text description $c_i$ as a condition, the motion generation model aims to generate a human motion sequence $m$ as $m=p^{1:L}$, where $L$ denotes the motion length. Each frame of the motion represents a pose $p^i$, comprising 3D joint rotations, positions, and velocities. 

In this paper, we adopt Motion Latent Diffusion (MLD) \cite{mld} as the basic model, denoted as $\hat{p}^{1:L}=\pi_{\rm mld}(c_i)$. For the basic motion generation process, MLD aims to model the conditional distribution of motions given a set of text description and motions $\{(c_i, p^{1:L}_i)\}$. In our work, our objective is to improve the generalization ability of motion generation models in unseen texts $\{c_j\}$ based on LLM reasoning and RL refinement. 

\subsection{Overview}
The overview framework of \ours{} is illustrated in Fig.~\ref{fig:overview}.
In \ours, we aim to use the strong reasoning capabilities and extensive knowledge of LLMs to convert a user-provided open-vocabulary textual description into executable motions. To achieve this, we decompose the task into three subtasks using a top-down approach, which are detailed in the following parts:
(i) \textbf{MCTS-enhanced LLM Reasoning}: The first step involves translating the textual description of the desired motion into the key joint positions of motion frames. This is accomplished through LLM-based reasoning enhanced by MCTS. (ii) \textbf{Motion completion and finetuning}: Building on the key joint positions, we estimate the full-body poses for the keyframes using a human pose model. Then the model generates smooth, fine-grained motions between the keyframes by finetuning a pre-trained MLD model with a dynamic temporal matching objective. (iii) \textbf{Physics-aware refinement}: We further adopt the MDP formulation of RL to post-train the learned model to address the physical implausibility arising from reasoning errors.

\subsection{MCTS-enhanced LLM Reasoning}

Existing LLMs generally lack spatial reasoning capabilities and physical knowledge \cite{jiang2024harmon,ji2025robobrain}, making it highly challenging to directly employ them for motion planning, since human motions involves spatial information such as 3D joint positions. Moreover, due to the strong correlations between frames in a motion sequence, generating a smooth and semantically meaningful motion demands that LLMs possess strong reasoning abilities.

To address this, we employ test-time scaling (i.e., MCTS) to enhance the reasoning ability of LLMs by progressively identifying promising keyframe prefixes. Specifically, for planning keyframes with a length of $K$, we construct a \emph{Motion Keyframe Tree (MKT)} to organize multiple planning traces, where each node $v$ stores the key joint positions $j_{\rm key}$ for a segment of $K_s$ keyframes. In MKT, a node at depth $d$ represents the $d$-th keyframe segment, and the path from the root node to a leaf node at depth $d$ corresponds to a complete keyframe sequence of length $K_s \cdot d$. The maximum depth of the MKT is $d_\text{max} = \lceil K/K_s \rceil$. 

In the reasoning process, the tree grows iteratively up to the $K$-th keyframe, where each sequence is rewarded based on the quality of the complete keyframe generation. The construction of the MKT follows the classic four stages of MCTS: \textbf{Selection}, \textbf{Expansion}, \textbf{Simulation}, and \textbf{Backpropagation}. Below, we describe each stage in detail.



\subsubsection{Selection} 
At each iteration, the algorithm traverses the tree from the root by selecting the child node with the maximum UCT value, until it reaches a leaf node:
\begin{equation}\label{eq1}
    v^* = \arg\max_{v' \in \text{Children}(v)} \Bigg[ Q(v') + \alpha \sqrt{\frac{2 \ln N(v)}{N(v')}} \Bigg],
\end{equation}
where $ Q(v') $ denotes the average reward of node $ v' $, $ N(v) $ and $ N(v') $ represent the visit counts of the parent and child nodes, respectively, and $\alpha$ is the exploration parameter.

\subsubsection{Expansion} 
If the selected is a leaf node, it is expanded by generating additional keyframe segments. The keyframe prefix for a leaf node $v$, is constructed by concatenating all keyframe segments along the path from the root to $v$. The LLM is then prompted to complete the keyframe sequence up to length $K$ using a predefined prompt template provided in Appendix A. The first keyframe segment produced in this completion is then stored as a child node for $v$.

\subsubsection{Simulation} 
Each node is rewarded by evaluating the quality of the complete keyframe sequence derived from the prefix defined by the node and its parent nodes. Specifically, for every complete keyframe sequence generated by the LLM, we reconstruct the full-body poses using a human pose model and render the corresponding sequence of keyframe images $\{I_i\}_{i=1}^K$. These rendered images are then evaluated for semantic consistency with the motion instruction text.
For consistency measurement, we use a CLIP-based evaluation model $\phi(\cdot)$, calculating the score as the average cosine similarity between text and rendered motion as:
\begin{equation}\label{eq2}
      \text{Score} = \frac{1}{K} \sum_{i=1}^{K} \cos\big( \phi_{\text{text}}(c), \phi_{\text{image}}(I_i) \big),
\end{equation}
where $\phi_{\text{text}}$ and $\phi_{\text{image}}$ denote the CLIP text encoder and CLIP image encoder, respectively.

\subsubsection{Backpropagation} 
The computed score is propagated back along the search path, updating the visit count and accumulated reward for each node as,
\begin{equation}
\label{eq:update}
  \forall u \in \text{Path(root} \rightarrow v_{\text{new}}): \quad
  \begin{cases}
     N(u) \gets N(u) + 1, \\
     W(u) \gets W(u) + \text{Score},
  \end{cases}
\end{equation}
where $ W(u) $ denotes the accumulated reward of node $ u $. Then, we update both the nodes in the path and their children as:
\begin{equation}\label{eq4}
    Q(u) = \frac{W(u)}{N(u)}, \quad 
    UCT(u) = Q(u) + \alpha \sqrt{\frac{2 \ln N(\text{parent}(u))}{N(u)}}.
\end{equation}
Additionally, if the selected node is a terminal node (the node reached maximum depth), only $N(u)$ is updated in Eq.~\eqref{eq:update} to encourage the exploration of alternative solution spaces. 

With MCTS-enhanced reasoning, \ours~ensures high-quality keyframe generation by systematically searching and selecting the best candidates from LLM-generated paths. This sets a solid foundation for subsequent full-body estimation and fine-grained temporal motion generation.


\subsection{Motion Completion and Finetuning}
\subsubsection{Full-body Pose Optimization}

For each keyframe, the LLM only plans the positions of the pelvis and four end-effector joints (denoted as $j_{\rm key}$), leaving the task of estimating a full-body pose. A naive idea would be to treat the target pose $\hat{p}$ as learnable parameters, compute joint positions via Forward Kinematics (FK), and optimize by minimizing the distance between the FK-derived joint positions and the LLM-specified target $j_{\rm key}$. However, this approach rarely yields reasonable poses due to insufficient constraints.

Thus, we leverage VPoser~\cite{SMPL-X:2019}, a learned human pose prior, to regularize the optimization space. While the original VPoser architecture comprises an encoder $E$ and a decoder $D$ that were trained on AMASS~\cite{AMASS:2019}, we further enhance its representational capacity by performing additional training on the Motion-X~\cite{lin2023motionx} dataset. This enables more complete coverage of the human pose space. Subsequently, instead of directly optimizing the target pose $\hat{p}$, we map it to the latent space as $\hat{z}=E(\hat{p})$ and treat $\hat{z}$ as learnable parameters. The pose is then reconstructed as $\hat{p}^{'}=D(\hat{z})$, from which we compute the full joint positions $\hat{j}= FK(\hat{p}^{'})$. Then, the optimization objective can be formulated as,
\begin{equation}\label{eq5}
    \mathcal{L}_{\rm pose}=\left\|M_j \cdot \hat{j}-j_{\rm key}\right\|+\left\|\hat{z}\right\|_{2}^{2},
\end{equation}
where $M_j\in\{0,1\}^{5\times22}$ is a joint mask exclusively selecting the planned joints, and $\left\|\hat{z}\right\|_{2}^{2}$ is a regularization term that penalizes deviations of $\hat{z}$ from the learned latent distribution. By minimizing $\mathcal{L}_{\rm pose}$, we obtain an optimal latent code $z^*$ such that the decoded pose $p^{*}$ simultaneously adheres to the human pose prior and satisfies the LLM-planned joint positions $j_{\rm key}$.

\subsubsection{Spatiotemporal Motion Completion}
Given a text prompt and a target motion length, through LLM planning and full-body pose optimization, we obtain key poses that represent the description. Subsequently, to generate a complete motion sequence, we finetune a pretrained MLD model using these incomplete key poses, leveraging MLD’s learned motion prior for temporal and spatial completion, while also enabling the model to adapt to new text-motion mapping. 

However, the planned key poses and prior-consistent motions are not inherently temporally aligned. To address this, we introduce a dynamic temporal matching objective $\mathcal{L}_{\mathrm{temporal}}$ based on soft-DTW~\cite{cuturi2017sdtw}, which allows flexible temporal alignment during fine-tuning. 
Specifically, given the key pose sequence $P^*=[p_1^*,\ldots,p_K^*]$ and the MLD-generated sequence $P=[p_1,\ldots,p_L]$ where $K<L$, we first construct a pairwise distance matrix $D\in\mathbb{R}^{K\times L}$, with each entry $D_{i,j}=\|p_i^*-p_j\|_2^2$ measuring the Euclidean distance between poses. We recursively calculate the cumulative cost matrix $R\in\mathbb{R}^{K\times L}$ as,
\begin{equation}\label{eq6}
    \begin{aligned}
        R_{i,j}&\!=\!D_{i,j}\!+\!\mathrm{min}^{\gamma}(R_{i-1,j},R_{i,j-1},R_{i-1,j-1})\\
        &\!\!\!\!=\! D_{i,j}\!-\!\underbrace{\gamma\log\left(e^{-R_{i-1,j}/\gamma}\!+\!e^{-R_{i,j-1}/\gamma}\!+\!e^{-R_{i-1,j-1}/\gamma}\right)}_{\text{soft minimum operator}},
    \end{aligned}
\end{equation}
where $\gamma > 0$ is a smoothing paramater. The $\mathcal{L}_{\mathrm{temporal}}$ is defined as $\mathcal{L}_{\mathrm{temporal}}=R_{K,L}$. Additionally, to ensure the precise adherence to key poses and prevent significant deviations, we augment the objective with a reconstruction loss:
\begin{equation}\label{eq7}
    \mathcal{L}_{\mathrm{recon}}=\frac{1}{K}\sum_{i=1}^{K}\|p_{i}^{*}-p_{\tau(i)}\|_{2}^{2},
\end{equation}
where $\tau(i)$ maps key pose's index to its aligned timestep in $P$. In practical implementations, MLD predicts the noise $\epsilon$, and we obtain the sequence $P$ through inverse reparameterization. The final optimization objective is formulated as,
\begin{equation}\label{eq8}
    \mathcal{L}_{\mathrm{MLD}} = \mathcal{L}_{\mathrm{recon}} + \lambda \cdot\mathcal{L}_{\mathrm{temporal}},
\end{equation}
where $\lambda$ is a weight paramater. By optimizing this joint objective, we strike a balance between temporal warping flexibility and spatial precision, achieving high-quality spatiotemporal motion completion.

\subsection{Physics-aware Refinement}

To address physical implausibility issues arising from reasoning errors, we implement RL post-training with physics-aware rewards to further refine motions. 

\subsubsection{MDP Formulation for RL Post-training}
Let $p_{\theta}(m_{0:T}|c)$ be the MLD model, where $c$ is some motion description distributed according to $p(c)$ and $m_{0:T}$ are motions in the MLD denoising steps. We frame the reverse denoising process as a Markov Decision Process $\mathcal{M}=(\mathcal{S}, \mathcal{A}, \mathcal{R}, \mathcal{P})$ followed by~\cite{fan2023dpok}:
\begin{itemize}
    \item State spaces $\mathcal{S}$: 
    The state $s_t$ at timestep $t$ is defined as $s_t=(c,m_{T-t})$ and $s_t \in \mathcal{S}$.
    \item Action spaces $\mathcal{A}$: The action at timestep $t$ is defined as $a_t = m_{T-t-1}$ and $a_t \in \mathcal{A}$.
    \item Reward function $\mathcal{R}$: Designed based on some motion physical plausibility rewards $r$ and only the final state $s_T$ receives a non-zero reward, which is determined by $m_0$, i.e.,
    \begin{equation}
        \mathcal{R}(s_t,a_t)=
        \begin{cases}
        r(s_{t+1})=r(m_0) & \mathrm{if} \ t=T-1, \\
        0 & \text{otherwise.}
        \end{cases}
    \end{equation}
    
    \item Transition function $\mathcal{P}$: The initial distribution $\mathcal{P}_0(s_0)=(p(c),\mathcal{N}(0,I))$ and the transition dynamics $P(s_{t+1}\mid s_t,a_t)=(\delta_c,\delta_{a_t})$, where $\delta$ is Dirac delta distribution.
\end{itemize}
Under this MDP, the denoising model acts as a parameterized policy $\pi_\theta(a_t\mid s_t) = p_\theta(m_{T-t-1}\mid m_{T-t},c)$ and initiates by sampling an initial state $s_{0}\sim \mathcal{P}_{0}$, which corresponds to sample Gaussian noise $m_T$ at the beginning of the reverse denoising process. We post-training this policy aims to maximize the expected reward of the generated motions.

\subsubsection{Physics-aware Reward Functions}
To enhance the physical plausibility of generated motions during post-training, we adopt three metrics proposed in \cite{han2025reindiffuse}: (i) The foot sliding reward imposes constraints by penalizing foot movements that deviate from expected locomotion patterns; (ii) The floating reward penalizes floating postures when foot-ground contact is required; (iii) The ground penetration reward prevents body parts from sinking below the ground plane. 





\subsubsection{Post-training Objective}
The objective of RL post-training is to maximize the expected physics-aware reward to refine the quality of generated motions, defined as:
\begin{equation}\label{eq10}
  J(\theta)=\mathbb{E}_{p_\theta(m_0|c)}[r(m_0)]
\end{equation}
We employ the Proximal Policy Optimization (PPO)~\cite{schulman2017ppo} algorithm to maximize the objective $J(\theta)$, leveraging importance sampling to enhance sample efficiency. This yields the following loss function:
\begin{equation}\label{eq11}
    \begin{aligned}
        &\mathcal{L}^{\mathrm{PPO}}(\theta)=\mathbb{E}_{p_{\theta_{\mathrm{old}}}(\boldsymbol{m}_{0:T}|c)}\Bigg[\sum_{t=1}^T-r(m_0)\cdot \\ 
        &\max\left(\rho_k(\theta,\theta_{\mathrm{old}}),\mathrm{clip}\left(\rho_k(\theta,\theta_{\mathrm{old}}),1+\epsilon\right)\right)\Bigg],
    \end{aligned}
\end{equation}
where $\rho_k(\theta,\theta_{\mathrm{old}}) = \frac{p_\theta(m_{t-1}|m_t,c)}{p_{\theta_{\mathrm{old}}}(m_{t-1}|m_t,c)}$ and $\epsilon$ is a clip hyperparameter. Additionally, we incorporate a Kullback-Leibler (KL) divergence objective, which can be defined as:
\begin{equation}\label{eq12}
    \mathcal{L}_{\mathrm{KL}}(\theta)=\mathbb{E}_{p_\theta}\left[\mathrm{KL}\left(p_\theta(m_{t-1}|m_t,c)\parallel p_{\mathrm{old}}(m_{t-1}|m_t,c)\right)\right]
\end{equation}
This objective can regularize policy updates, preventing excessive deviation from the original policy.
\section{Experiments}

In this section, we present experiments to evaluate the effectiveness of \ours{}. Our experiments aim to answer the following three research questions:
\begin{itemize}
    \item \textbf{Q1.} Can \ours{} generate reasonable motions for open-vocabulary descriptions?
    \item \textbf{Q2.} Do MCTS and dynamic temporal matching improve motion generation performance?
    \item \textbf{Q3.} Does physics-aware refinement improve motions' physical plausibility in practical scenarios?
\end{itemize}

\begin{table}[!h]\centering
\caption{Quantitative comparison of the quality of generated motions.} \label{tab:main-result}
\resizebox{\linewidth}{!}{ 
\begin{tabular}{lcccc}\hline
Methods       & CLIP\_S $\uparrow$& VLM\_S $\uparrow$& Float $\downarrow$& Pene $\downarrow$\\ \hline
MDM          & 21.77\textsubscript{$\pm$0.78}      & 1.12\textsubscript{$\pm$0.16}        & 26.98\textsubscript{$\pm$16.12}    & 31.49\textsubscript{$\pm$2.76}       \\
MLD          & 21.38\textsubscript{$\pm$0.79}         & 1.44\textsubscript{$\pm$0.22}        & \textbf{2.10\textsubscript{$\pm$8.6}}    & 38.55\textsubscript{$\pm$2.15}       \\
MotionGPT    & 20.27\textsubscript{$\pm$0.73}      & 1.06\textsubscript{$\pm$0.24}        & 10.75\textsubscript{$\pm$8.81}    & 37.59\textsubscript{$\pm$1.58}       \\
MotionCLIP   & 21.54\textsubscript{$\pm$0.77}         & 0.75\textsubscript{$\pm$0.13}       & 24.17\textsubscript{$\pm$3.82}       & 50.28\textsubscript{$\pm$7.45}          \\
AnySkill     & 20.04\textsubscript{$\pm$0.92}         & 0.95\textsubscript{$\pm$0.17}        & -        & -           \\ \hline
Ours(w/o RL) & \textbf{24.03\textsubscript{$\pm$0.45}}      & 2.69\textsubscript{$\pm$0.18}        & 16.85\textsubscript{$\pm$10.01}    & 32.84\textsubscript{$\pm$2.47}       \\
Ours(full)   & 23.64\textsubscript{$\pm$0.49}      & \textbf{2.72\textsubscript{$\pm$0.21}}        & 2.46\textsubscript{$\pm$1.05}       & \textbf{21.70\textsubscript{$\pm$2.70}}          \\ \hline
\end{tabular}
}
\vspace{-1em}
\end{table}

\begin{figure*}
    \centering
    \includegraphics[width=0.90 \linewidth]{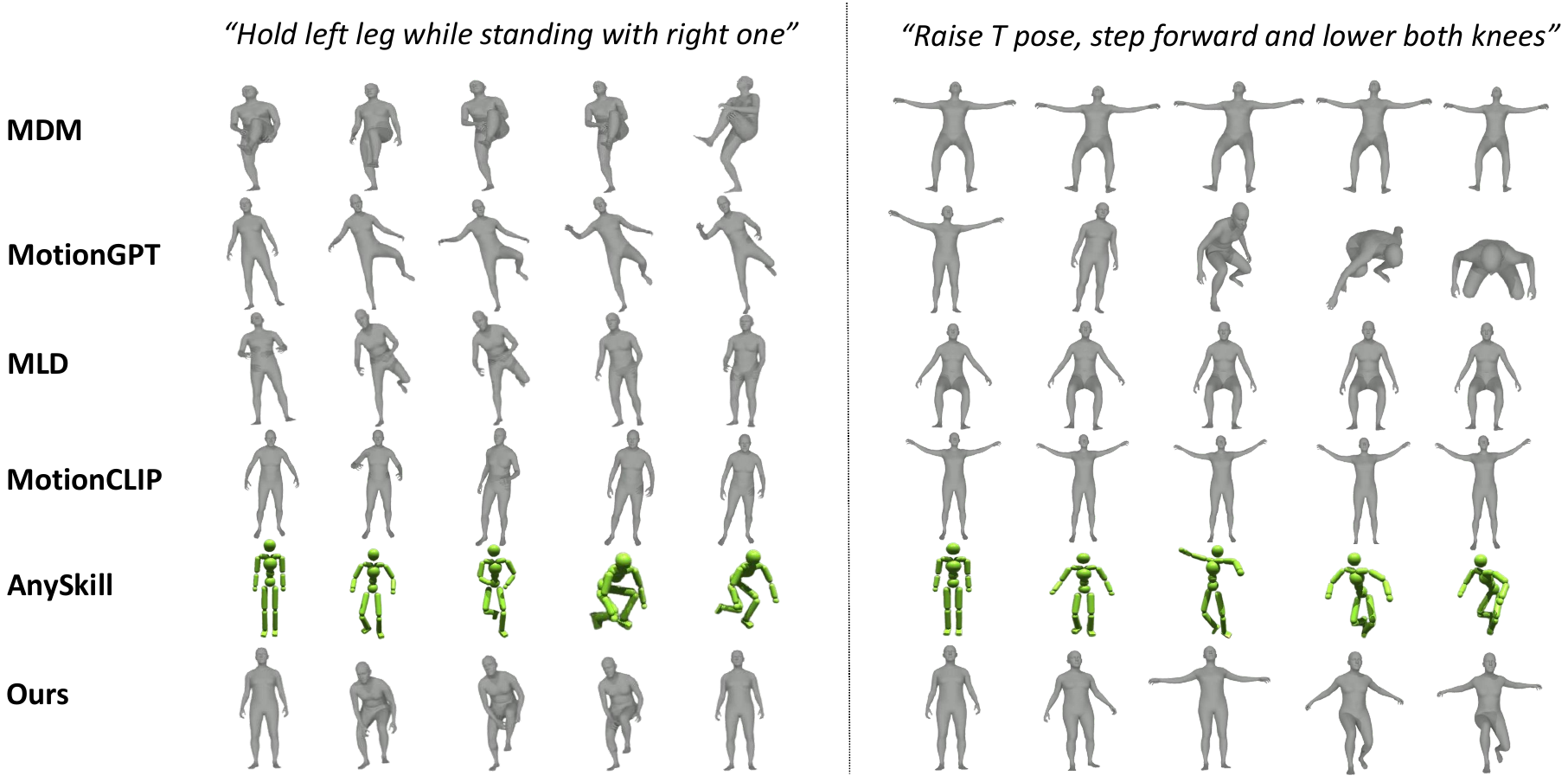}
    \caption{Qualitative comparison results of motions generated by different methods.}
    \label{fig:experiment}
\end{figure*}

\subsection{Experiment Setup}
\subsubsection{Dataset}

Our experiments are conducted on multiple pose and text-motion paired datasets, including HumanML3D~\cite{CVPR2022hml3d}, AMASS~\cite{AMASS:2019}, and Motion-X~\cite{lin2023motionx}. (i) HumanML3D is a widely used motion language dataset. In our experimental setup, we reconstruct it by partitioning out $40$ texts as the unseen texts and compute the CLIP text similarity between unseen texts and the remaining texts in the dataset. We filter out remaining pairs with similarity greater than a preset threshold, e.g., $0.75$, resulting in a text-motion paired subset. This subset is used to pre-train the MLD model employed in our motion completion stage. Moreover, for those $40$ unseen texts, we remove the subject part, e.g., ``a person bends down and jumps up'' is transformed into ``bend down and jump up,'' making it more aligned with the open-vocabulary instruction style. The processed texts are then used for open-vocabulary motion generation experiments. (ii) AMASS is an unlabeled human motion dataset that consolidates various optical marker-based motion capture datasets into a standardized format. Motion-X is a large-scale motion dataset containing expressive full-body human motions paired with corresponding text labels. The original VPoser model is trained on AMASS, and we perform additional training on Motion-X without text labels to improve its capability for human pose representation. 

\subsubsection{Evaluation Baselines}
We compare our method with several baselines, including MDM~\cite{mdm}, MLD~\cite{mld}, MotionGPT~\cite{jiang2024motiongpt}, MotionCLIP~\cite{tevet2022motionclip}, and AnySkill~\cite{cui2024anyskill}. Both MDM and MLD are diffusion-based methods that exhibit excellent performance in fitting the dataset, demonstrating a certain degree of generalization capability. MotionGPT compresses the motion and text spaces using discrete tokens, which allows it to generalize well to novel text combinations. MotionCLIP maps the motion space to the CLIP text space, leveraging CLIP's generalization ability, thus enabling it to handle open-vocabulary tasks to some extent. AnySkill utilizes the exploration capability of an agent in RL, combining with the CLIP similarity as a reward signal, thereby guiding the agent to explore actions that align with the open-vocabulary descriptions.

\subsubsection{Evaluation Metrics}
We evaluate the generated motions from two aspects: semantic alignment and physical plausibility. For motion semantic alignment, we use CLIP score (\textbf{CLIP\_S}) and VLM score (\textbf{VLM\_S}) as quantitative evaluation metrics. Specifically, we render the generated motions to videos and compute the average CLIP similarity score between each frame of the videos and the corresponding text descriptions. Additionally, we input the videos into a visual language model (VLM), i.e., QWen-VL, to score the motion based on semantic alignment and naturalness, with the weighted score serving as the VLM score. For physical plausibility, we mainly follow the \textbf{Floating} and \textbf{Penetration} metrics from PhysDiff to measure the degree of floating and ground penetration (all in mm). The detailed definition of metrics is given in Appendix B.

\begin{table}[ht]\centering
\caption{Ablation study of our method.} \label{tab:ablation}
\begin{tabular}{lcc} \hline
Methods     & CLIP\_S$\uparrow$     & VLM\_S$\uparrow$   \\ \hline
w/o MCTS    & 22.57\textsubscript{$\pm$0.41}        & 1.96\textsubscript{$\pm$0.26}       \\
w/o $L_{\text{temporal}}$ & 23.16\textsubscript{$\pm$0.54} & 1.57\textsubscript{$\pm$0.11} \\ \hline
Ours        & \textbf{23.64\textsubscript{$\pm$0.49}}        & \textbf{2.72\textsubscript{$\pm$0.21}}  \\ \hline
\end{tabular}
\vspace{-1em}
\end{table}

\begin{table*}[ht]
    \caption{Ablation of Physics-aware Refinement in Isaacgym.}\label{tab:rl-gym}
    \centering
    \begin{tabular}{lcccccc}
    \toprule
Methods&  $E_{\mathrm{mpbpe}} \downarrow$ & $E_{\mathrm{mpjpe}} \downarrow$ & $E_{\mathrm{mpbve}} \downarrow$ & $E_{\mathrm{mpbae}} \downarrow$ & $E_{\mathrm{mpjve}} \downarrow$ \\
    \midrule
Ours(w/o RL)   & 68.20\textsubscript{$\pm$17.25} & 1327.25\textsubscript{$\pm$127.24} & 
        10.08\textsubscript{$\pm$2.55} &
        10.63\textsubscript{$\pm$2.72} &  114.04\textsubscript{$\pm$15.75}  \\
Ours(full)    & \textbf{51.21\textsubscript{$\pm$12.22}} & \textbf{1086.69\textsubscript{$\pm$94.49}} & 
    \textbf{7.16\textsubscript{$\pm$1.82}}  &
    \textbf{7.45\textsubscript{$\pm$1.86}} & 
     \textbf{94.04\textsubscript{$\pm$13.74}} \\
    \bottomrule
    \end{tabular}
    \vspace{-1em}
\end{table*}

\subsection{Main Result}

To address \textbf{Q1}, we conduct a comprehensive evaluation comparing our approach with existing methods, divided into two groups: (i) models trained on the dataset and directly perform zero-shot generalization to unseen texts (e.g., MDM~\cite{mdm}, MLD~\cite{mld}, MotionCLIP~\cite{tevet2022motionclip}, MotionGPT~\cite{jiang2024motiongpt}), and (ii) methods with specific processing for unseen texts (e.g., AnySkill~\cite{cui2024anyskill}).

When comparing with direct evaluation methods, we train MDM, MLD, and MotionGPT on our reconstructed subset of HumanML3D. Since MotionCLIP requires frame-level annotations, we train it on a processed version of BABEL with the corresponding indexed entries removed. All methods are evaluated on those 40 unseen texts. Quantitative results are shown in Table~\ref{tab:main-result}. Conventional methods, e.g., MDM, MLD, and MotionGPT, which primarily fit the training distribution, exhibit limited generalization on open-vocabulary texts, resulting in poor semantic alignment. Moreover, MotionCLIP often generates discontinuous motions, leading to low scores in VLM-based evaluation. In contrast, our \ours{} achieves superior performance, owing to its LLM-based keyframe planning that leverages advanced spatial reasoning to generate robust initial motions for unseen text descriptions.

When comparing with AnySkill, we note it trains a dedicated high-level controller for each unseen text, using CLIP similarity as a reward signal to guide the agent toward text-aligned states. However, as shown in Table~\ref{tab:main-result}, AnySkill's performance is suboptimal, due to two key limitations: (i) the inherent instability of RL-based exploration, and (ii) its reward function that only evaluates immediate state transitions rather than complete motion sequences. While both methods employ specific processing for individual unseen text, our \ours{} achieves superior performance by leveraging LLM-based spatial reasoning capabilities. 

In terms of physical plausibility, since AnySkill is trained in simulation, we exclude its physical plausibility metrics from evaluation. It can be observed that RL post-training effectively mitigates physical implausibility arising from LLM reasoning errors.

Qualitative results for the generated motions are shown in Fig.~\ref{fig:experiment}. Since AnySkill is trained using the humanoid character from IsaacGym, its visual mesh differs from those of other methods. For the description `Hold left leg while standing with right one', the motion generated by MDM successfully raises the left leg; however, the hand does not `hold the left leg' but instead rests in front of the abdomen. Other methods generate motions that deviate from the description. Our \ours{}~generates a motion that accurately captures the textual description: raising the left leg, holding it with the hand, standing on the right leg, and finally returning to the initial position. For the description 'Raise T-pose, step forward, and lower both knees', most existing methods only capture the 'raise T-pose' while failing to generate the subsequent motions. Conversely, our approach effectively generates the complete motion sequence, demonstrating its superior capability in handling sequential descriptions. More qualitative results are shown in Appendix C.

\subsection{Impact of MCTS and Dynamic Temporal Matching}

To answer \textbf{Q2}, we conduct ablation studies in the same experimental settings to evaluate the impact of using MCTS and dynamic temporal matching, i.e., $\mathcal{L}_{\mathrm{temporal}}$. The results are in Table~\ref {tab:ablation}. Disabling MCTS restricts the LLM to a single planning attempt, limiting its ability to thoroughly explore the solution space and producing plans inconsistent with the input description. When Dynamic Temporal Matching is not applied and the model is fine-tuned solely with $\mathcal{L}_{\mathrm{recon}}$ as the objective, it forcibly aligns the generated motion with the keyframes planned by the LLM at the corresponding indices, resulting in unnatural motion sequences. This ablation study demonstrates that enhancing the LLM's reasoning with MCTS improves the likelihood of generating text-consistent plans and that dynamic temporal matching is essential for flexible temporal alignment.

\subsection{Impact of Physics-aware Refinement}
To answer \textbf{Q3}, we sample $5$ motions corresponding to the same textual descriptions before and after refinement, and train imitation policies to track these motions. We employ PBHC~\cite{xie2025kungfubot} as the policy learning framework, which represents the current state-of-the-art open-source methods. All policies are trained in IsaacGym, with each motion trained under $3$ random seeds to ensure statistical robustness.
The performance is evaluated with the following metrics: Global Mean Per Body Position Error ($E_\textrm{g-mpbpe}$, mm), root-relative Mean Per Body Position Error ($E_\textrm{mpbpe}$, mm), Mean Per Joint Position Error ($E_\textrm{mpjpe}$, $10^{-3}$ rad), Mean Per Joint Velocity Error ($E_\textrm{mpjve}$, $10^{-3}$ rad/frame), Mean Per Body Velocity Error ($E_\textrm{mpbve}$, mm/frame), and Mean Per Body Acceleration Error ($E_\textrm{mpbae}$, mm/frame\textsuperscript{2}).

The quantitative results are shown in Table~\ref {tab:rl-gym}, and the tracking accuracy of the policy significantly improves after physics-aware refinement. This suggests that the refined motions mitigate physically implausible artifacts, enabling the agent to imitate them more effectively.


\begin{figure}
    \centering
    \includegraphics[width=0.92 \linewidth]{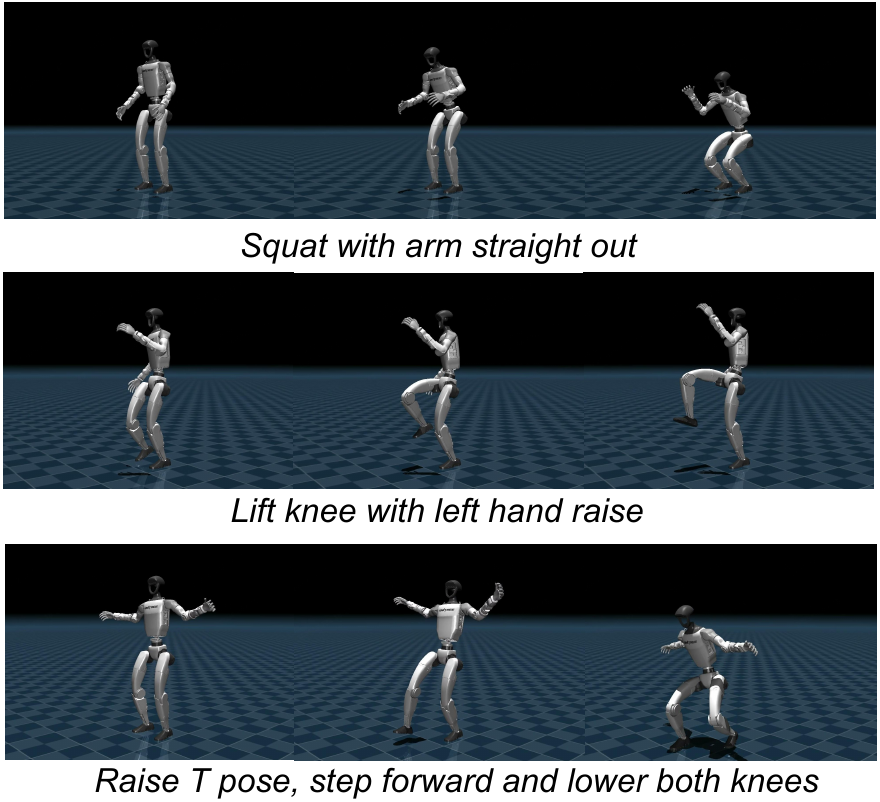}
    \caption{Visualization of our generated motions on the MuJoCo platform.}
    \label{fig:mujoco}
    \vspace{-1.5em}
\end{figure}

Furthermore, we conduct a sim-to-sim transfer of certain strategies on the MuJoCo platform, which track the physics-aware refinement motions. The visualization results are shown in Fig.~\ref{fig:mujoco}, and more results are in the Appendix section. Beyond simulation, Fig.~\ref{fig:robot} exhibits the successful deployment of these motions on physical hardware. This showcases the practical viability and real-world potential of our method.

\begin{figure}
    \centering
    \includegraphics[width=0.92 \linewidth]{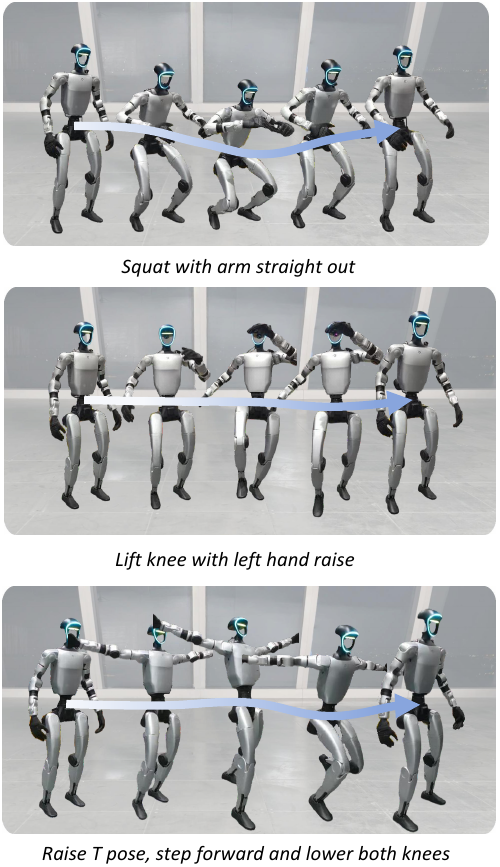}
    \caption{Deploy generated motions on the real world robot.}
    \label{fig:robot}
    \vspace{-1.5em}
\end{figure}

\section{Conclusion}
This paper proposes \ours, which leverages LLM reasoning and physics-aware refinement for generalizable motion generation in open-vocabulary. Our three-stage approach first employs MCTS-enhanced LLM reasoning to generate reasonable keyframes of the motion based on text prompts. Then, a human prior pose model is used to optimize full-body poses for these keyframes, and dynamic temporal matching based on soft-DTW completes the sequence spatiotemporally. Finally, RL post-training with physics-aware rewards refines physical plausibility. Extensive experiments demonstrate that \ours~can achieve superior performance in generating semantically aligned and physically plausible motions compared to existing methods.

\section{Acknowledgement}
We are grateful to Weiji Xie, Jinrui Han, Yuhang Lin, Jipeng Kong and Jiyuan Shi for their assistance with the real-robot experiments.

This work was supported by the National Natural Science Foundation of China (No.~62306115) and the independent research project of the State Key Laboratory of Spatial Intelligent Manipulation Technology.
{
    \small
    \bibliographystyle{ieeenat_fullname}
    \bibliography{main}
}

\clearpage
\setcounter{page}{1}
\maketitlesupplementary

\section{A. Implementation Details} \label{app:1}

\subsection{Hyperparameter}
In the LLM planning, we employ \textbf{DeepSeek-R1} as the reasoning model, which exhibits superior spatial understanding and reasoning capabilities. The CLIP model mentioned in the paper uses the \textbf{CLIP-ViT-L/14} model from OpenCLIP~\cite{cherti2023openclip}. The VLM used in our experiments is \textbf{Qwen-VL-Max}. The detailed hyperparameters involved in our method are presented in Table~\ref{tab:parameter}.

\begin{table}[h]
    \caption{Hyperparameters of \ours.}
    \label{tab:parameter}
    \centering
    \renewcommand{\arraystretch}{1.2}
    {\fontsize{10}{10}\selectfont
    \begin{tabular}{lc}
    \hline
        {Hyperparameter} & {Value } \\ \hline
        $K_s$ & 2 \\ 
        MCTS Iteration & 30 \\
        Exploration Parameter $\alpha$ & 0.05 \\ 
        Smoothing Parameter $\gamma$  & 0.1 \\ 
        Weight Parameter $\lambda$  & 0.01 \\ 
        MLD Fine-tune Batch Size & 64 \\
        Learning Rate & $1e-4$ \\
        PPO Clip Threshold & $1e-3$ \\ 
        Buffer Size & 3000 \\
        Samples Per Update Iteration & 8 \\
        Policy Training Batch Size & 128 \\
        KL Weight & 0.01 \\ \hline
    \end{tabular}}
\end{table}
\subsection{Prompt Design}
\begin{figure}[ht]
    \centering
    \includegraphics[width=0.15 \linewidth]{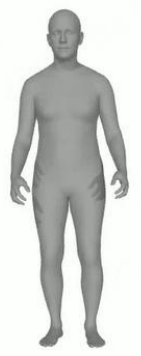}
    \caption{The initial pose for LLM planning.}
    \label{fig:initial_pose}
\end{figure}

\begin{figure*}[t]
    \centering
    \includegraphics[width=0.99 \linewidth]{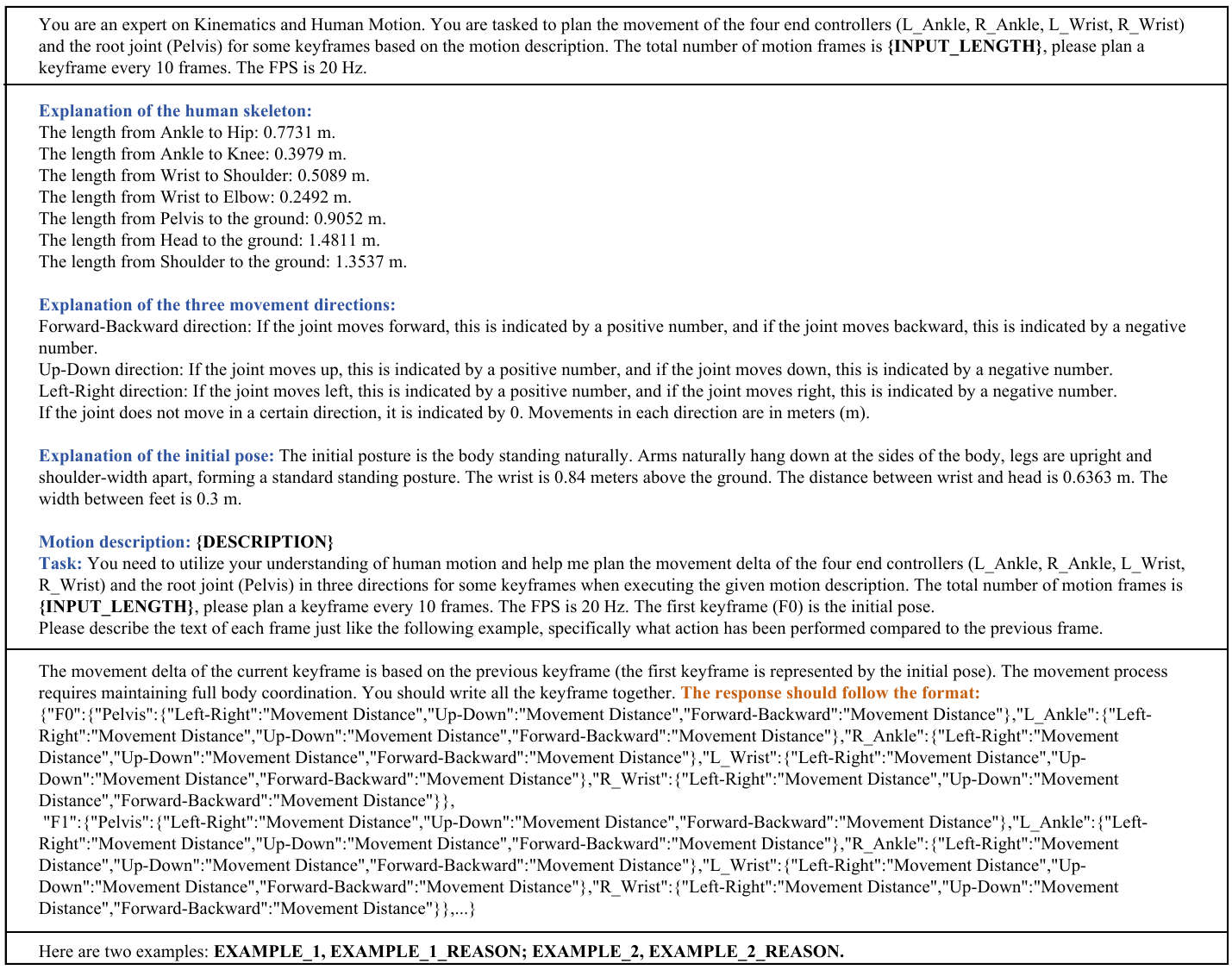}
    \caption{Our prompt template for LLM reasoning.}
    \label{fig:main_prompt}
\end{figure*}

\begin{figure*}[t]
    \centering
    \includegraphics[width=0.99 \linewidth]{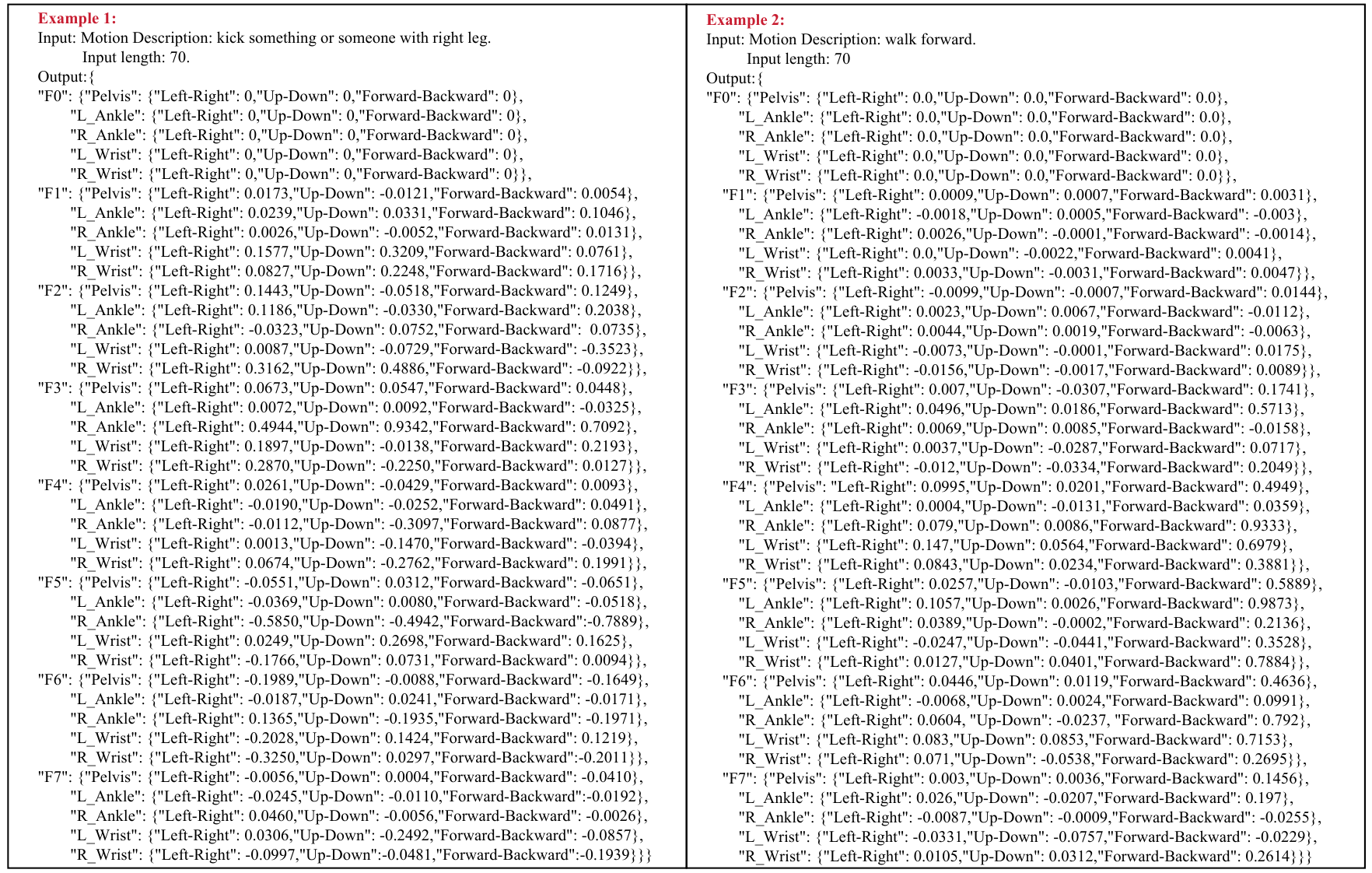}
    \caption{Examples for LLM reasoning.}
    \label{fig:example}
\end{figure*}

\begin{figure*}[t]
    \centering
    \includegraphics[width=0.99\linewidth]{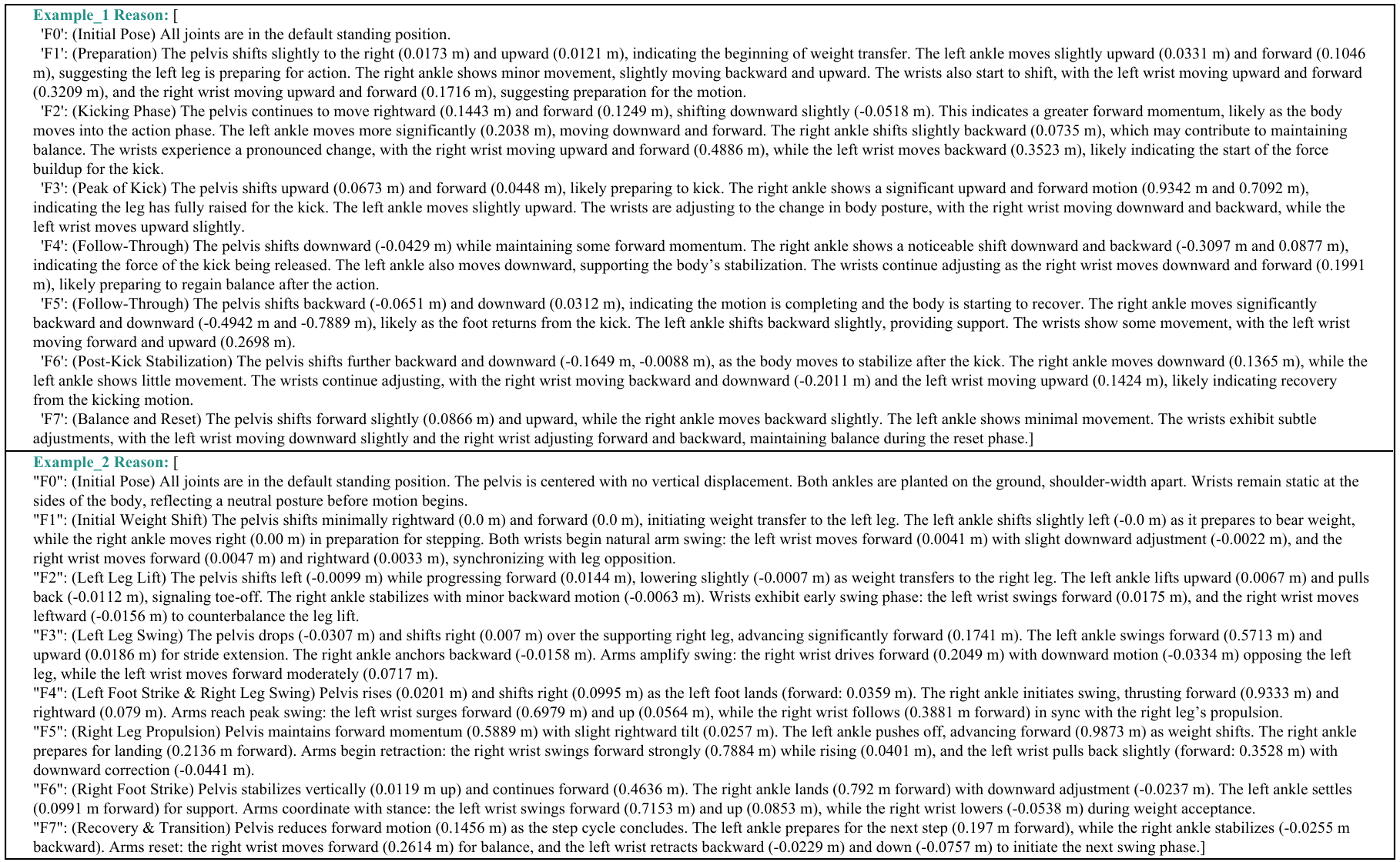}
    \caption{Related reasons for the given examples.}
    \label{fig:reason}
\end{figure*}

We use LLMs to plan keyframes for human motions. The quality of the planned keyframes depends not only on the model itself but also on the input prompt. Our prompt template is shown in Fig.~\ref{fig:main_prompt}, which includes four key points: 
\begin{enumerate}
    \item To simplify the reasoning task, we do not ask the LLM to plan motions for full-body joints. The reasons are: 1) excessive data processing would significantly degrade the reasoning quality of the model; 2) the strong inter-dependencies among full-body joints make the reasoning less tolerant to errors. Therefore, we only make the LLM infer motions for $5$ key joints and plan only the displacement of each keyframe relative to the previous one.
    \item We define a set of foundational information as follows:
    \begin{itemize}
        \item Human skeleton information, derived from the SMPL~\cite{loper2023smpl} Neutral skeleton.
        \item The displacement direction information is not represented in XYZ coordinates, but in directional terms such as ``Forward-Backward'', ``Up-Down'', and ``Left-Right'', which helps the LLM better grasp movement orientation.
        \item The initial body pose is shown in Fig.~\ref{fig:initial_pose}, which helps standardize the planning’s initialization conditions.
        \item Task information, reminding the LLM of its objective.
    \end{itemize}
    \item Additionally, we require the LLM to output the reasoning behind its planned keyframes, encouraging more thorough deliberation during inference. We provide a formatted output template to facilitate information extraction and subsequent usage. 
    \item We include two examples to achieve few-shot fine-tuning, enabling the LLM to quickly adapt to our task during inference. Fig.~\ref{fig:example} shows these two examples, and Fig.~\ref{fig:reason} explains the reasons.        
\end{enumerate}

With the above prompt template, the LLM's capability in planning such a task can be significantly enhanced.

\subsection{Motion Representations}

In our implementation, we utilize three distinct motion data representations:
\begin{enumerate}
    \item \textbf{During LLM reasoning}, motion is represented by $K$ planned keyframes. Each keyframe $j_{\text{key}}$ contains X-Y-Z coordinates for $5$ key joints (pelvis, l/r\_ankle, l/r\_wrist).
    
    \item \textbf{During full-body poses optimization}, the optimized poses follow the SMPL-format~\cite{loper2023smpl}, structured as a dictionary containing three components:
        \begin{itemize}
            \item Global translation: Root joint X-Y-Z coordinates representing global body position.
            \item Root rotation: Euler angles representing overall body rotation.
            \item Body pose: Euler angles for $21$ joints (excluding hands and facial joints).
        \end{itemize}   
    \item \textbf{During MLD fine-tuning and post-training}, the pose of each frame for the model input uses the 263-dimensional format defined by HumanML3D, derived from SMPL-format data. This 263-dimensional format of data includes root angular/linear velocity, root height, joint rotation invariant position, joint rotation, joint linear velocity, and foot contact information. It must be noted that we maintain consistent pose notation with full-body pose optimization for clarity in the main text, and that all HumanML3D-format data can be converted to/from SMPL-format. 
\end{enumerate}

\subsection{Physics-aware Reward Design}

In the physics-aware refinement section, we use RL post-training with the following three reward functions:
\begin{itemize}
    \item \textbf{Foot Sliding Reward.} This reward function imposes constraints by penalizing foot movements that deviate from expected locomotion patterns, formally defined as:
    \begin{equation}
        r_\mathrm{S}(m)=\frac{1}{L} \sum_{i=1}^{L} \exp(-\|(p_{ft}^i-p_{ft}^{i-1})\cdot p_c^i\cdot p_c^{i-1}\|_2),
    \end{equation}
    where $m=p^{1:L}$ is mentioned in the main text, $p_{ft}^i$ and $p_c^i$ represent the $i$-th pose's foot joint positions and the contact label, respectively.

    \item \textbf{Foot Floating Reward.} This reward function penalizes floating postures when foot-ground contact is required, which is formally defined as:
    \begin{equation}
        \begin{aligned}
            r_\mathrm{F}(m) &=\frac{1}{L} \sum_{i=1}^{L} \exp(-||(\operatorname*{min}_{v}p_{i}^{v}-h_{ground}) \\
            & \cdot \mathbb{I}[\operatorname*{min}_{v}p_{i}^{v}>h_{\text{ground}}]||_2),
        \end{aligned}
    \end{equation}
    where $\operatorname*{min}_{v}p_{i}^{v}$ represents the lowest body mesh vertex of the pose $p_i$, $h_{ground}$ denotes the height of ground.

    \item \textbf{Ground Penetration Reward.} This reward function encourages all motions remain grounded, defined as:
    \begin{equation}
        \begin{aligned}
            r_{\mathrm{P}}(m)&=\frac{1}{L} \sum_{i=1}^{L} \exp(-\|(h_{ground}-\operatorname*{min}_{v}p_{i}^{v})\\
            & \cdot \mathbb{I}[\operatorname*{min}_{v}p_{i}^{v}<h_{\text{ground}}]\|_2).
        \end{aligned}
    \end{equation}
\end{itemize}

\section{B. Experiments Details}\label{app:2}
\subsection{LLM-planned Keyframes}

\begin{figure*}[t]
    \centering
    \includegraphics[width=0.99 \linewidth]{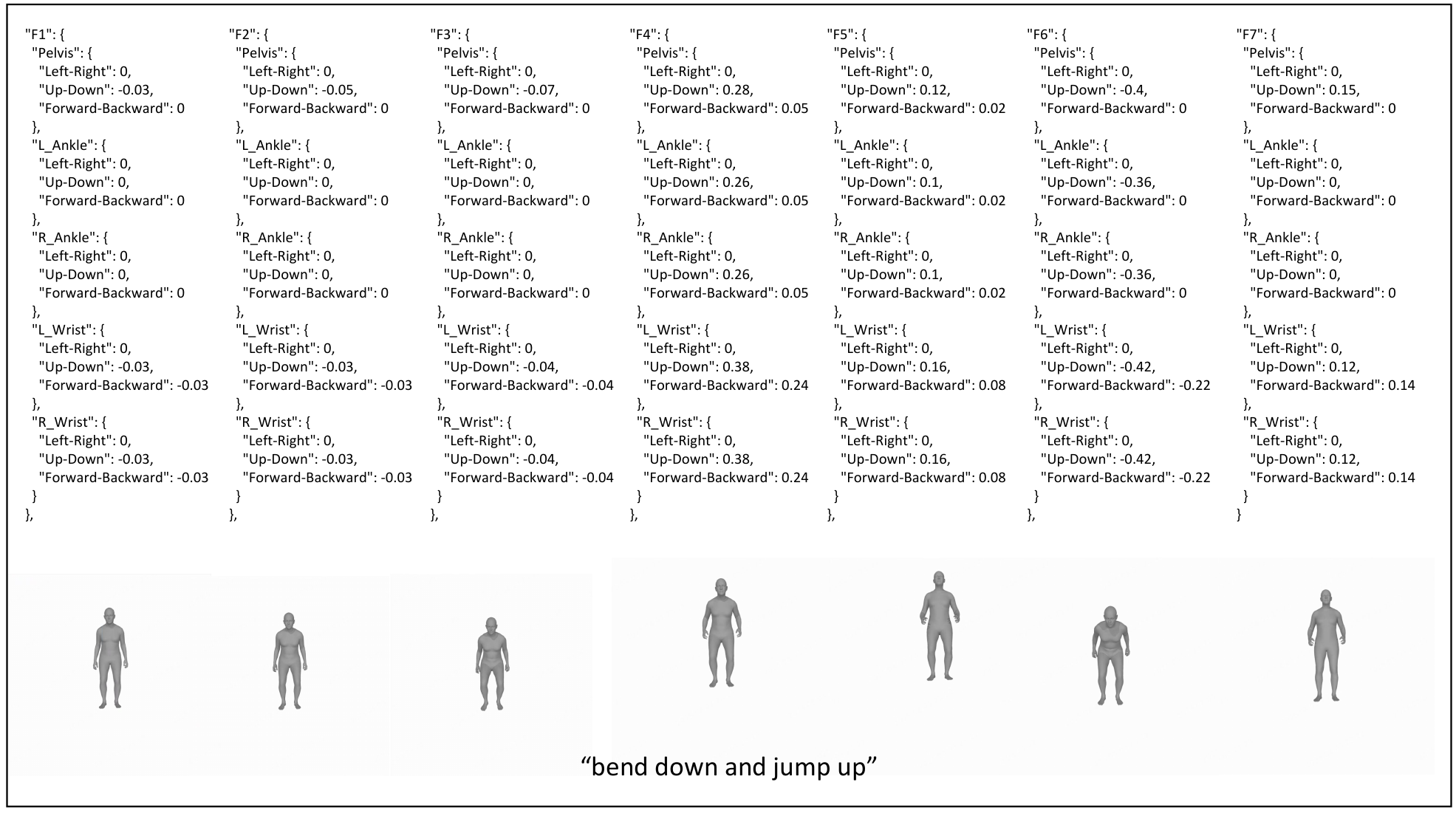}
    \caption{LLM-planned key joint positions and rendered images.}
    \label{fig:keyframe1}
\end{figure*}

\begin{figure*}[t]
    \centering
    \includegraphics[width=0.99 \linewidth]{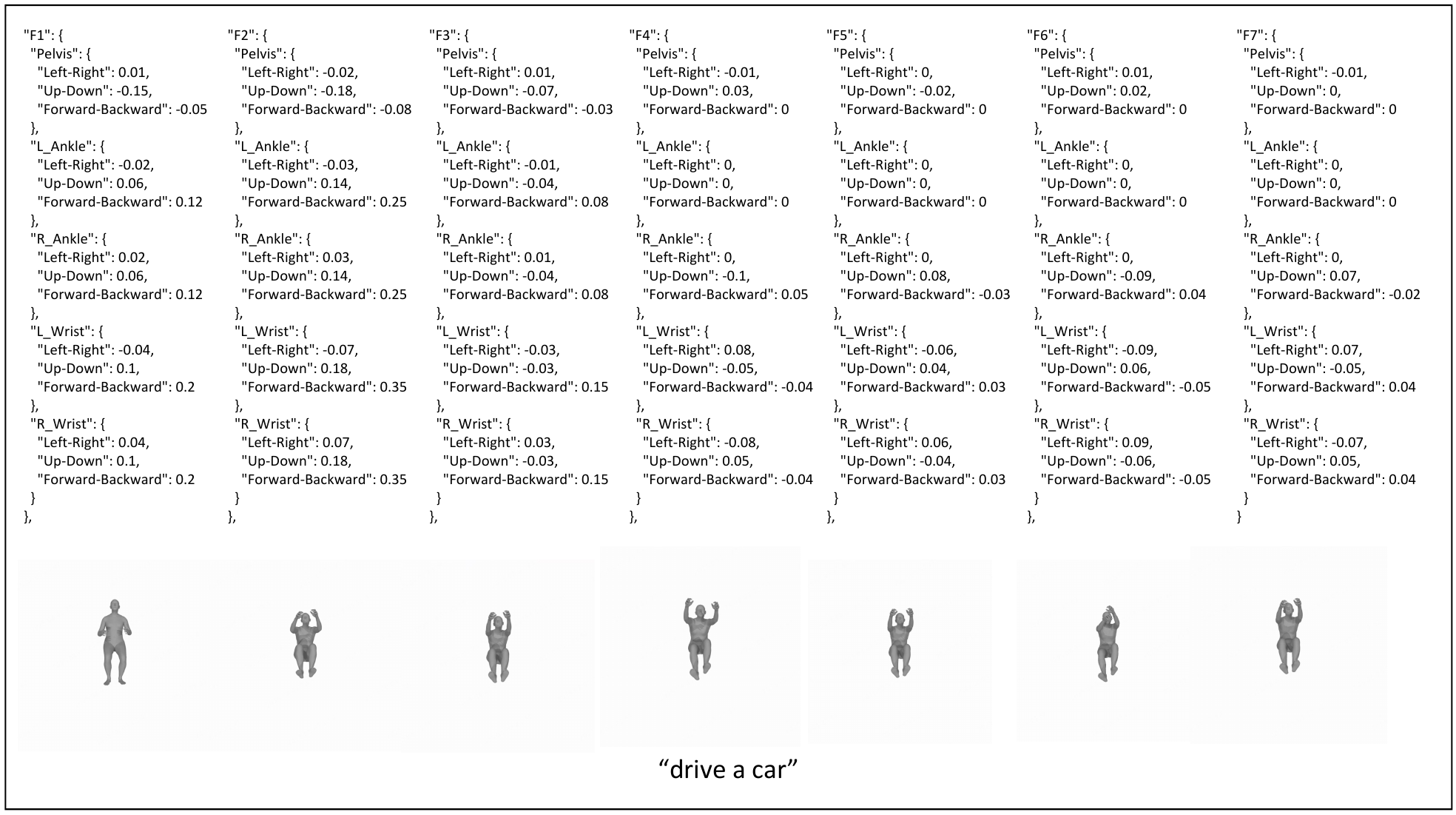}
    \caption{LLM-planned key joint positions and rendered images.}
    \label{fig:keyframe2}
\end{figure*}

\begin{figure*}[t]
    \centering
    \includegraphics[width=0.99 \linewidth]{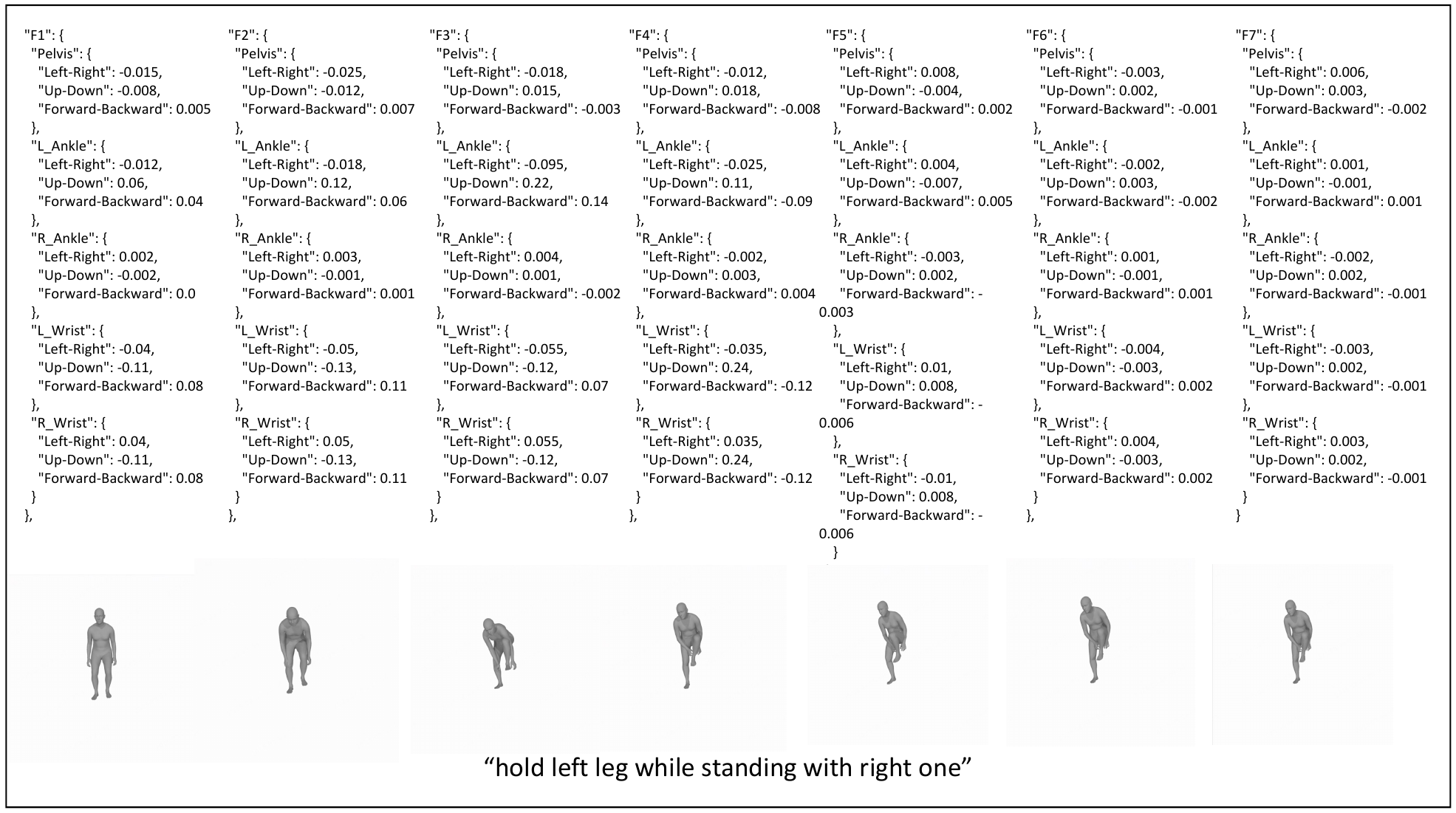}
    \caption{LLM-planned key joint positions and rendered images.}
    \label{fig:keyframe3}
\end{figure*}

\begin{figure*}[t]
    \centering
    \includegraphics[width=0.99 \linewidth]{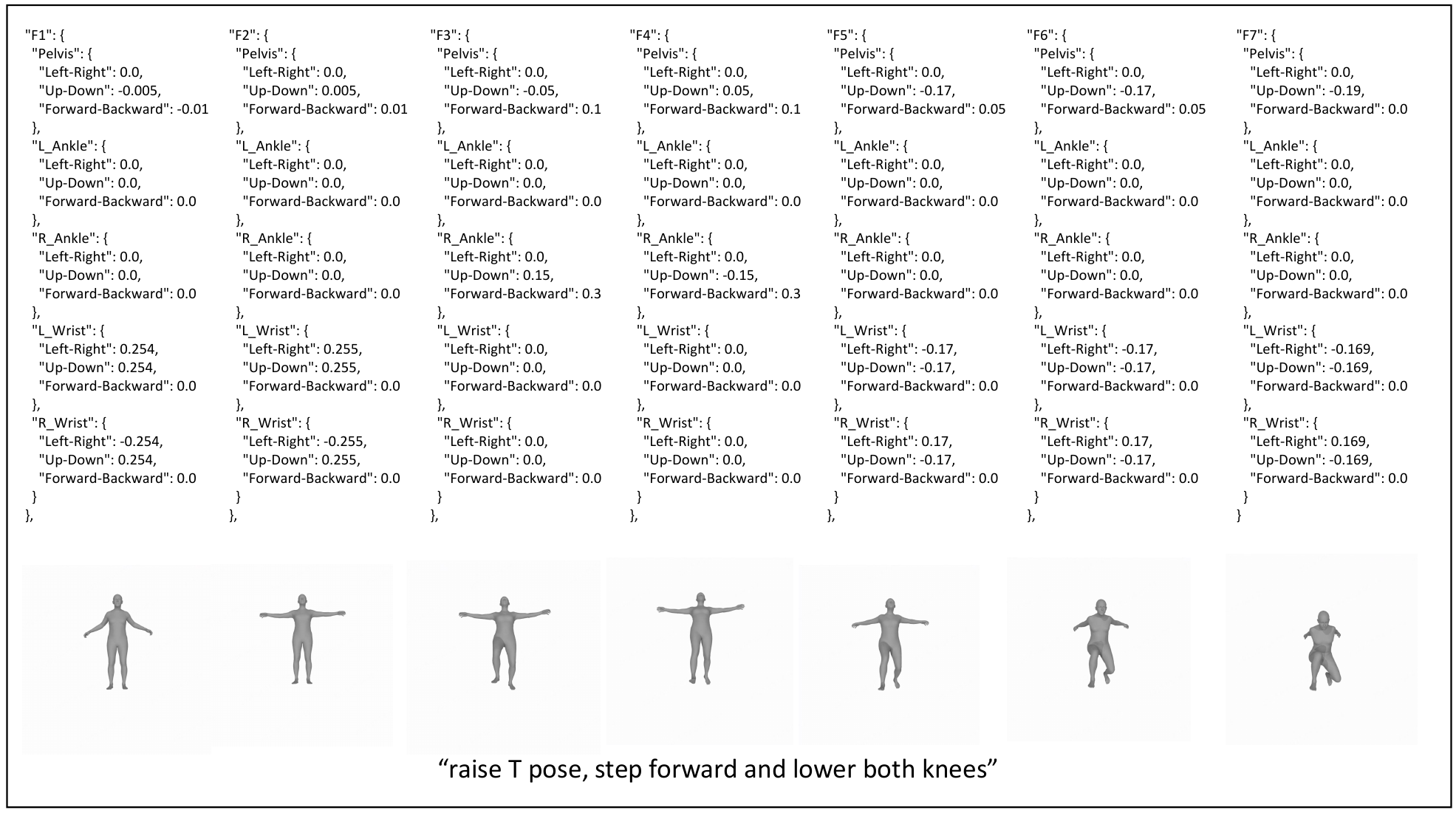}
    \caption{LLM-planned key joint positions and rendered images.}
    \label{fig:keyframe4}
\end{figure*}

\begin{figure*}[t]
    \centering
    \includegraphics[width=0.99 \linewidth]{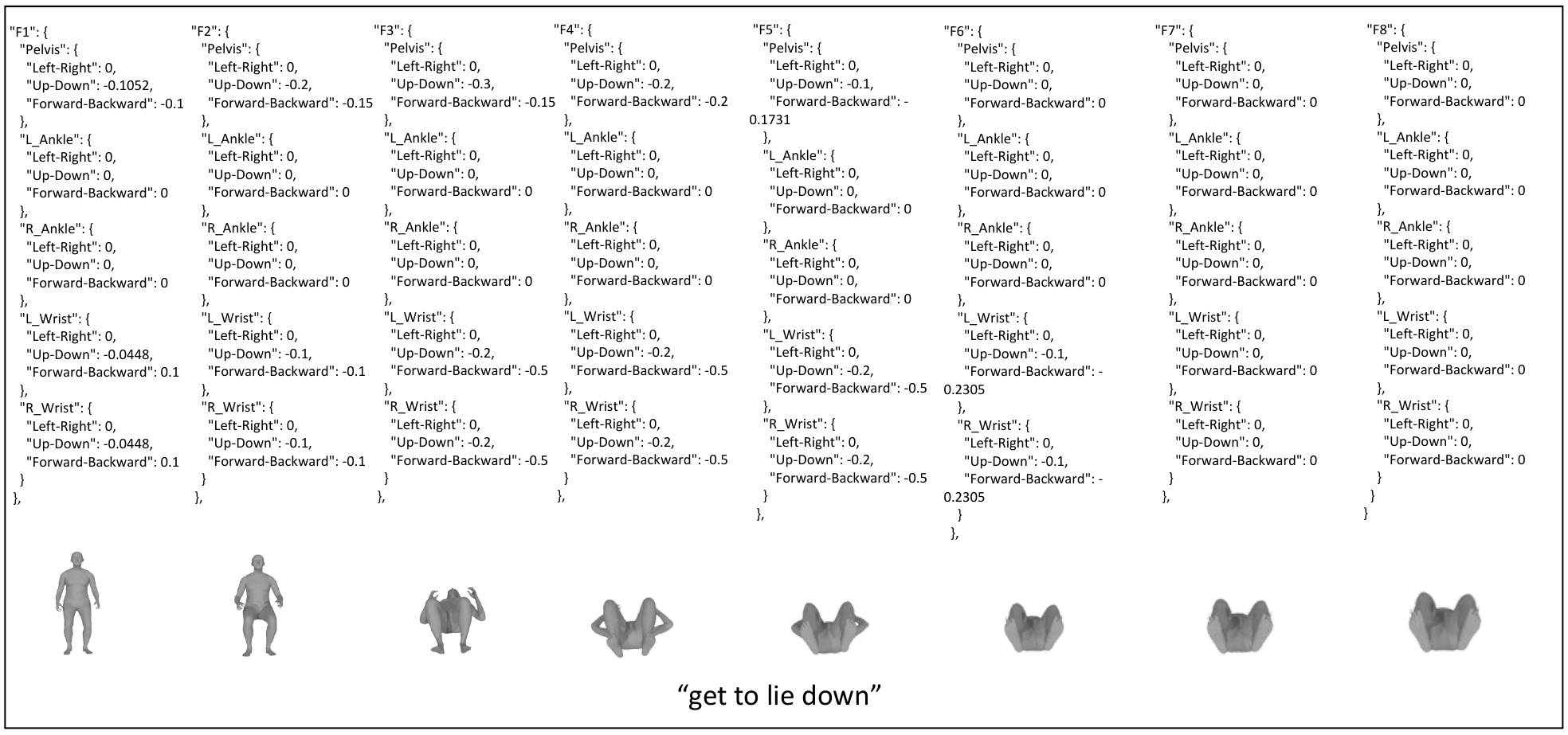}
    \caption{LLM-planned key joint positions and rendered images.}
    \label{fig:keyframe5}
\end{figure*}

\begin{figure*}[t]
    \centering
    \includegraphics[width=0.99 \linewidth]{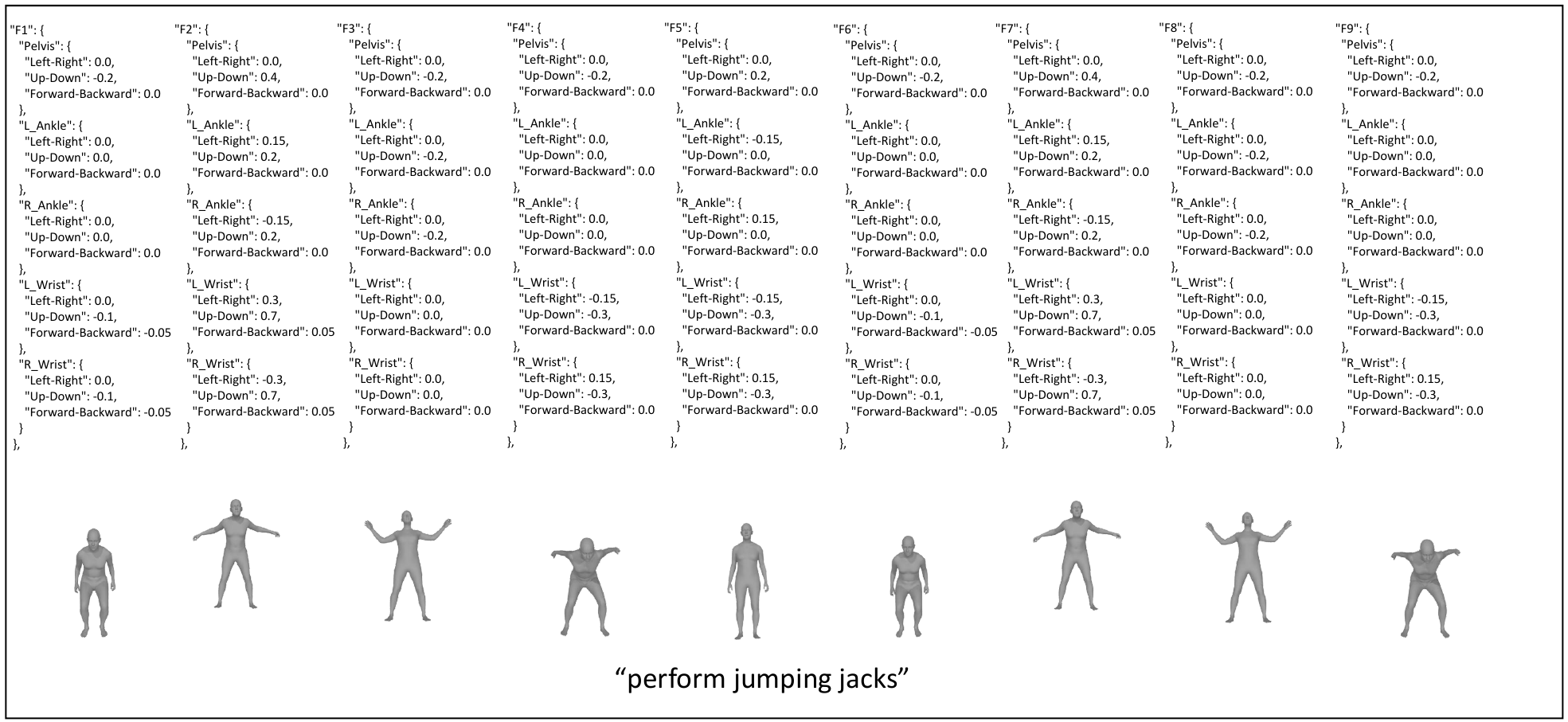}
    \caption{LLM-planned key joint positions and rendered images.}
    \label{fig:keyframe6}
\end{figure*}

\begin{figure*}[t]
    \centering
    \includegraphics[width=0.99 \linewidth]{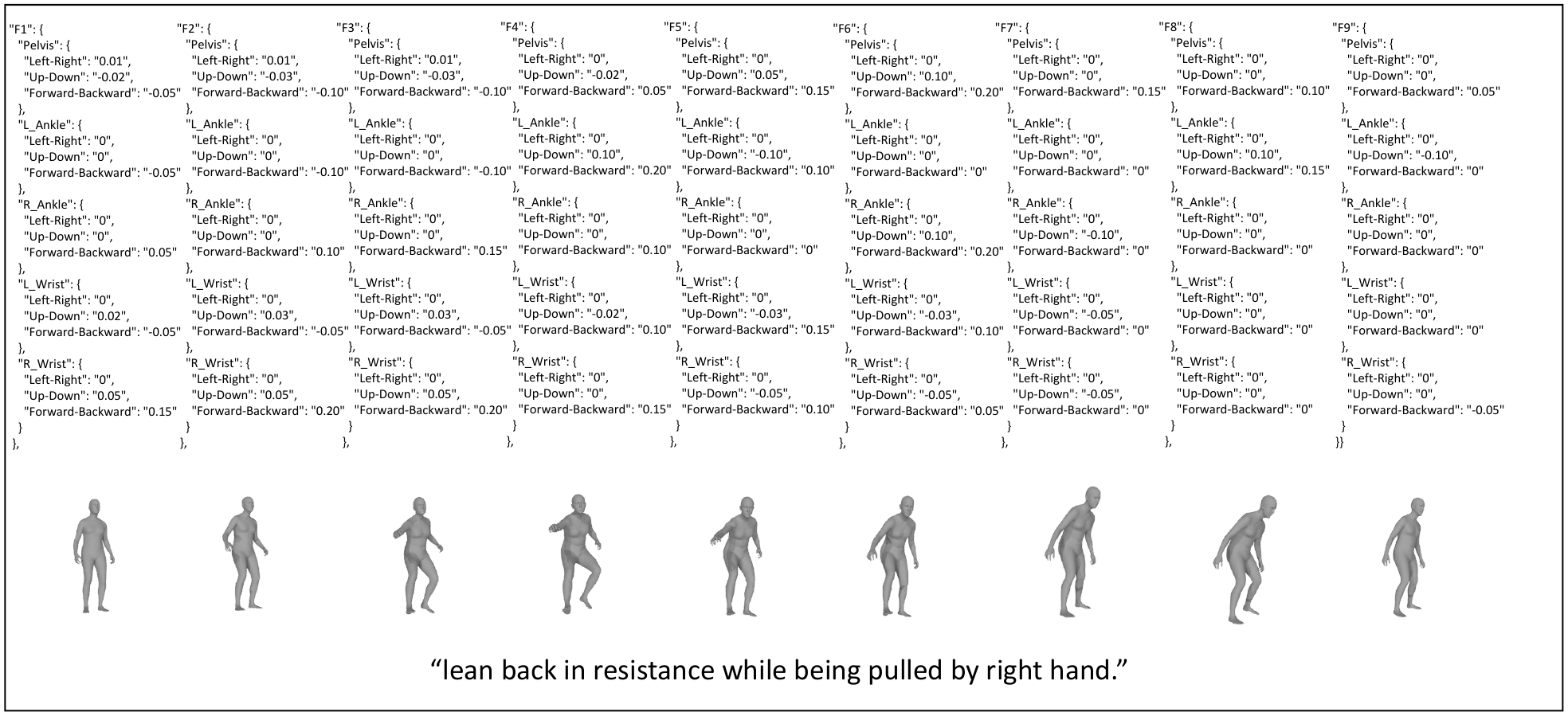}
    \caption{LLM-planned key joint positions and rendered images.}
    \label{fig:keyframe7}
\end{figure*}

As shown in Figs.(\ref{fig:keyframe1}-\ref{fig:keyframe7}), we present the JSON data of keyframes planned by the LLM of different motion lengths alongside the rendered pose images after full-body optimization. These results demonstrate that LLM can reasonably plan actions based on text descriptions.

\begin{figure*}[t]
    \centering
    \includegraphics[width=0.92 \linewidth]{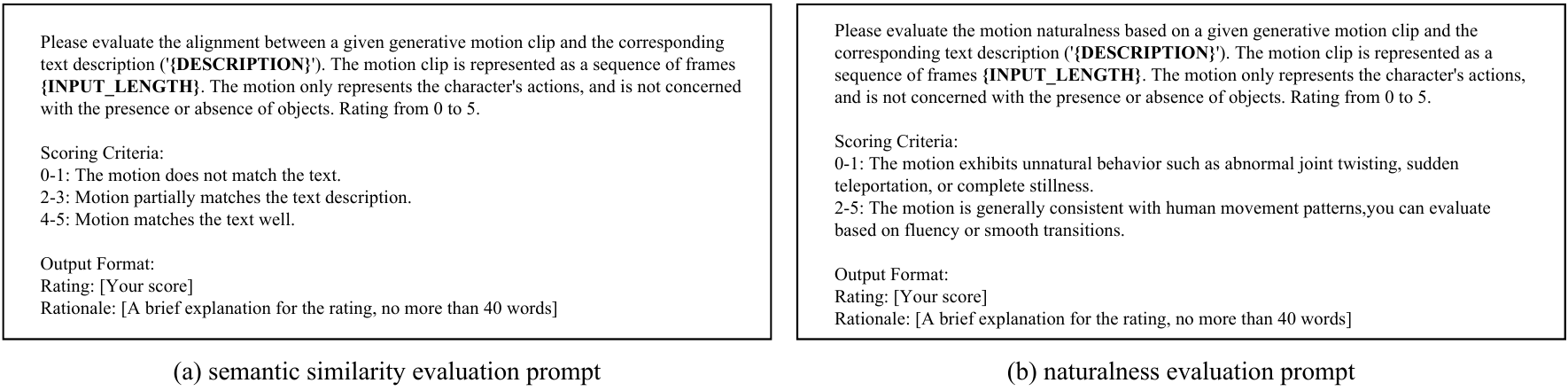}
    \caption{VLM evaluation prompt.}
    \label{fig:vlm_prompt}
\end{figure*}

\subsection{Evaluation Metrics}

As mentioned in the main text, we evaluate the generated motions from two aspects: semantic alignment and physical plausibility. Thus, we use the following four metrics:
\begin{itemize}
    \item CLIP score (CLIP\_S): The average clip similarity between each frame of the rendered motion videos and the corresponding description, which is calculated as,
    \begin{equation}
        \text{CLIP\_S} = \frac{1}{L} \sum_{i=1}^{L} \cos\big( \phi_{\text{text}}(c), \phi_{\text{image}}(I_i) \big),
    \end{equation}
    where $L$ is the motion length, $c$ is the motion description, $I_i$ represents the rendered image of each motion frame. $\phi_{\text{text}}$ and $\phi_{\text{image}}$ denote the CLIP text encoder and CLIP image encoder, respectively. 
    
    \item VLM score (VLM\_S): The weighted score combines the VLM's evaluations of semantic alignment and naturalness for the rendered motion video $v$, denoted as,
    \begin{equation}
        \text{VLM\_S} = \sigma_s \cdot \text{VLM}_{\text{S}}(v,c) + \sigma_n \cdot \text{VLM}_{\text{N}}(v),
    \end{equation}
    where $\text{VLM}_{\text{S}}$ and $\text{VLM}_{\text{N}}$ are distinct evaluation prompts designed to assess semantic alignment and naturalness, respectively, as illustrated in Fig.~\ref{fig:vlm_prompt}. The weights $\sigma_s$ and $\sigma_n$ are set to $0.6$ and $0.4$ respectively. For each motion, we compute the average of 10 VLM scores to ensure statistical robustness.

    \item Floating (Float): The degree of floating from the ground.
    \begin{equation}
        \text{Float} = \frac{1}{L}\sum_{i=1}^{L}\lvert \operatorname*{min}_{v}p_{i}^{v} - h_{\text{ground}}\rvert\cdot\mathbb{I}[\operatorname*{min}_{v}p_{i}^{v}>h_{\text{ground}}],
    \end{equation}
    where $\operatorname*{min}_{v}p_{i}^{v}$ represents the lowest body mesh vertex of the pose $p_i$.

    \item Penetration (Pene): The degree of ground penetration.
        \begin{equation}
        \text{Pene} = \frac{1}{L}\sum_{i=1}^{L}\lvert \operatorname*{min}_{v}p_{i}^{v} - h_{\text{ground}}\rvert\cdot\mathbb{I}[\operatorname*{min}_{v}p_{i}^{v}<h_{\text{ground}}].
    \end{equation}
\end{itemize}

\section{C. Additional Experiments}\label{app:3}  

\begin{table}[ht]
\centering
\small
\caption{Results on SnapMotion dataset.} 
\label{tab:snap}
\begin{tabular}{lcc} 
\toprule
Methods      & CLIP\_S $\uparrow$ & VLM\_S $\uparrow$ \\ 
\midrule
MLD & $23.39\scriptstyle{\pm 0.40}$ & $1.64\scriptstyle{\pm 0.24}$ \\ 
MotionGPT & $22.36\scriptstyle{\pm 0.51}$ & $1.42\scriptstyle{\pm 0.14}$ \\ 
MoMask & $23.92\scriptstyle{\pm 0.36}$ & $1.75\scriptstyle{\pm 0.17}$ \\ 
\midrule
Ours & $\mathbf{24.68\scriptstyle{\pm 0.64}}$ & $\mathbf{2.17\scriptstyle{\pm 0.40}}$ \\ 
\bottomrule
\end{tabular}
\vspace{-1em}
\end{table}

\subsection{Experiments on SnapMotion}
Table~\ref{tab:snap} demonstrates the performance of our method on $100$ text descriptions from the SnapMotion~\cite{snapmogen2025} dataset, comparing it with methods in motion generation, including MoMask~\cite{guo2024momask}. The results highlight the superior capability of our approach in generating motions from unseen text descriptions.

\subsection{Impact of Different LLMs}

To evaluate the impact of different LLMs on the experimental results, we also conduct experiments on three other LLMs, i.e, Qwen-Plus, InternLM, and DeepSeek-V3. 
Qwen-Plus and DeepSeek-R1, used in our experiments, possess deep thinking capabilities, while InternLM and DeepSeek-V3 do not. The result is shown in Table~\ref{tab:abl_llm}. It can be observed that LLMs with deep thinking capabilities demonstrate superior performance, as they can more thoroughly analyze tasks, reflect on the rationality of their planning and reasoning steps, and correct their thought processes. Among these, the DeepSeek-R1 model we employed exhibits particularly outstanding abilities in this regard.

\begin{table}[ht]\centering
\caption{Ablation study with different LLMs.} \label{tab:abl_llm}
\begin{tabular}{lcc} \hline
Methods     & CLIP\_S$\uparrow$     & VLM\_S$\uparrow$   \\ \hline
DeepSeek-V3    & 22.59\textsubscript{$\pm$0.75}        & 1.77\textsubscript{$\pm$0.16}       \\
InternLM & 22.1\textsubscript{$\pm$0.67} & 1.54\textsubscript{$\pm$0.25} \\ 
Qwen-Plus & 23.12\textsubscript{$\pm$0.51} & 1.83\textsubscript{$\pm$0.24} \\ \hline
DeepSeek-R1 (Ours)        & \textbf{23.64\textsubscript{$\pm$0.49}}        & \textbf{2.72\textsubscript{$\pm$0.21}}  \\ \hline
\end{tabular}
\end{table}

\subsection{Analysis of Smoothing Parameter $\gamma$}

To analyze the effect of different smoothing parameter $\gamma$ values, we conduct additional experiments with $\gamma$ = \{0.01, 0.05, 0.5, 1\}. The results are shown in Table~\ref {tab:abl_gamma}. It can be observed that our method remains robust across a wide range of $\gamma$ values, and we ultimately set $\gamma$ = 0.1 as the final choice.

\begin{table}[ht]\centering
\caption{Results of different smoothing parameter values.} \label{tab:abl_gamma}
\begin{tabular}{lcc}\hline
    $\gamma$          & \multicolumn{1}{c}{CLIP\_S$\uparrow$}     & \multicolumn{1}{c}{VLM\_S$\uparrow$}     \\ \hline
$\gamma=0.01$ & \multicolumn{1}{c}{23.30\textsubscript{$\pm$0.79}} & \multicolumn{1}{c}{2.38\textsubscript{$\pm$0.32}} \\
$\gamma=0.05$ & 23.52\textsubscript{$\pm$0.52}       & 2.65\textsubscript{$\pm$0.27}   \\
$\gamma=0.5$  & \multicolumn{1}{c}{\textbf{23.68\textsubscript{$\pm$0.78}}} & \multicolumn{1}{c}{2.51\textsubscript{$\pm$0.18}} \\
$\gamma=1$    & 23.66\textsubscript{$\pm$0.82}     & 2.70\textsubscript{$\pm$0.26}   \\ \hline
$\gamma=0.1$ (Ours)  & 23.64\textsubscript{$\pm$0.49}   & \textbf{2.72\textsubscript{$\pm$0.21}}  \\ \hline
\end{tabular}
\end{table}

\begin{figure*}[t]
    \centering
    \includegraphics[width=0.99 \linewidth]{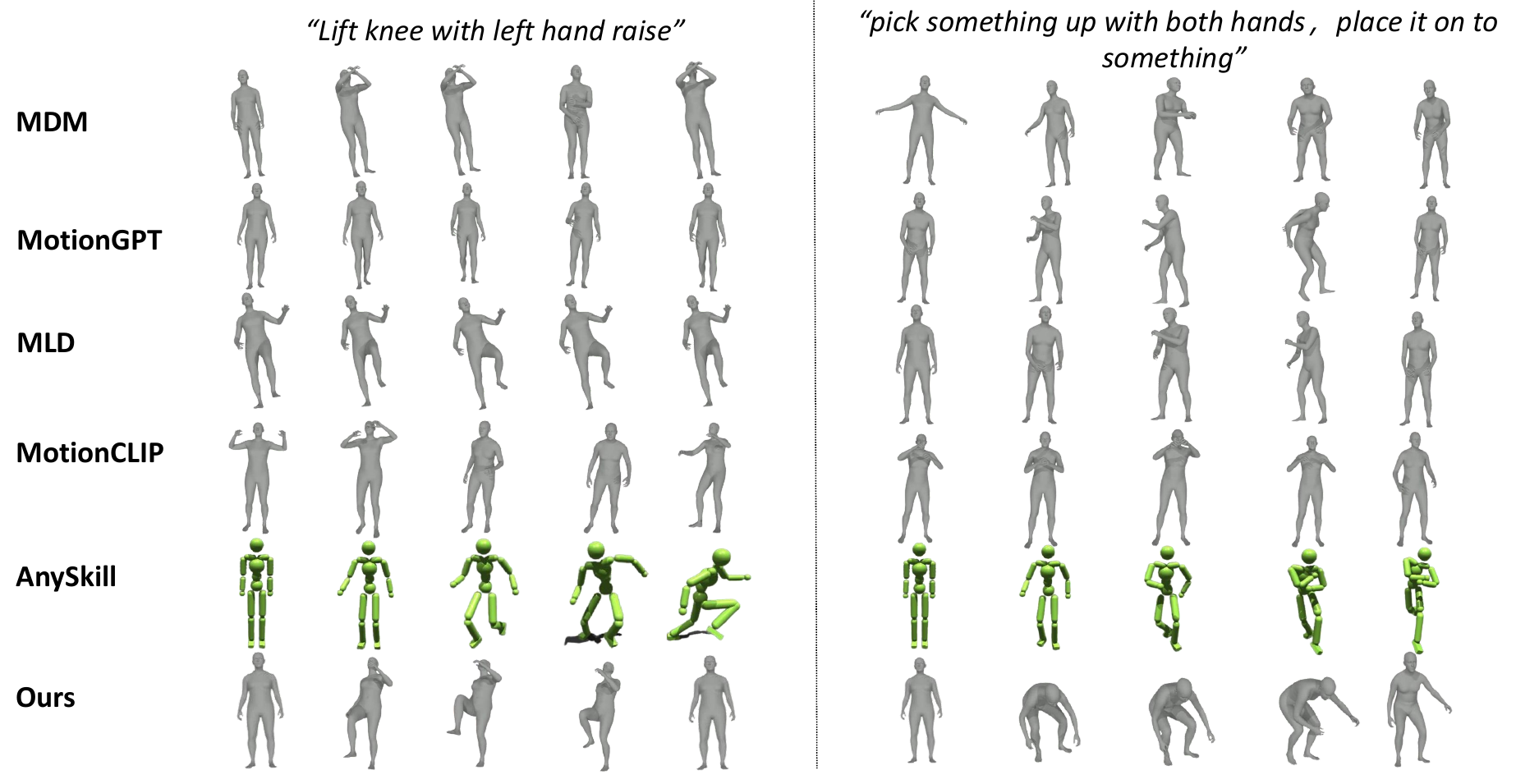}
    \caption{Additional qualitative comparison results of motions generated by different methods.}
    \label{fig:add_res_1}
\end{figure*}

\begin{figure*}[t]
    \centering
    \includegraphics[width=0.99 \linewidth]{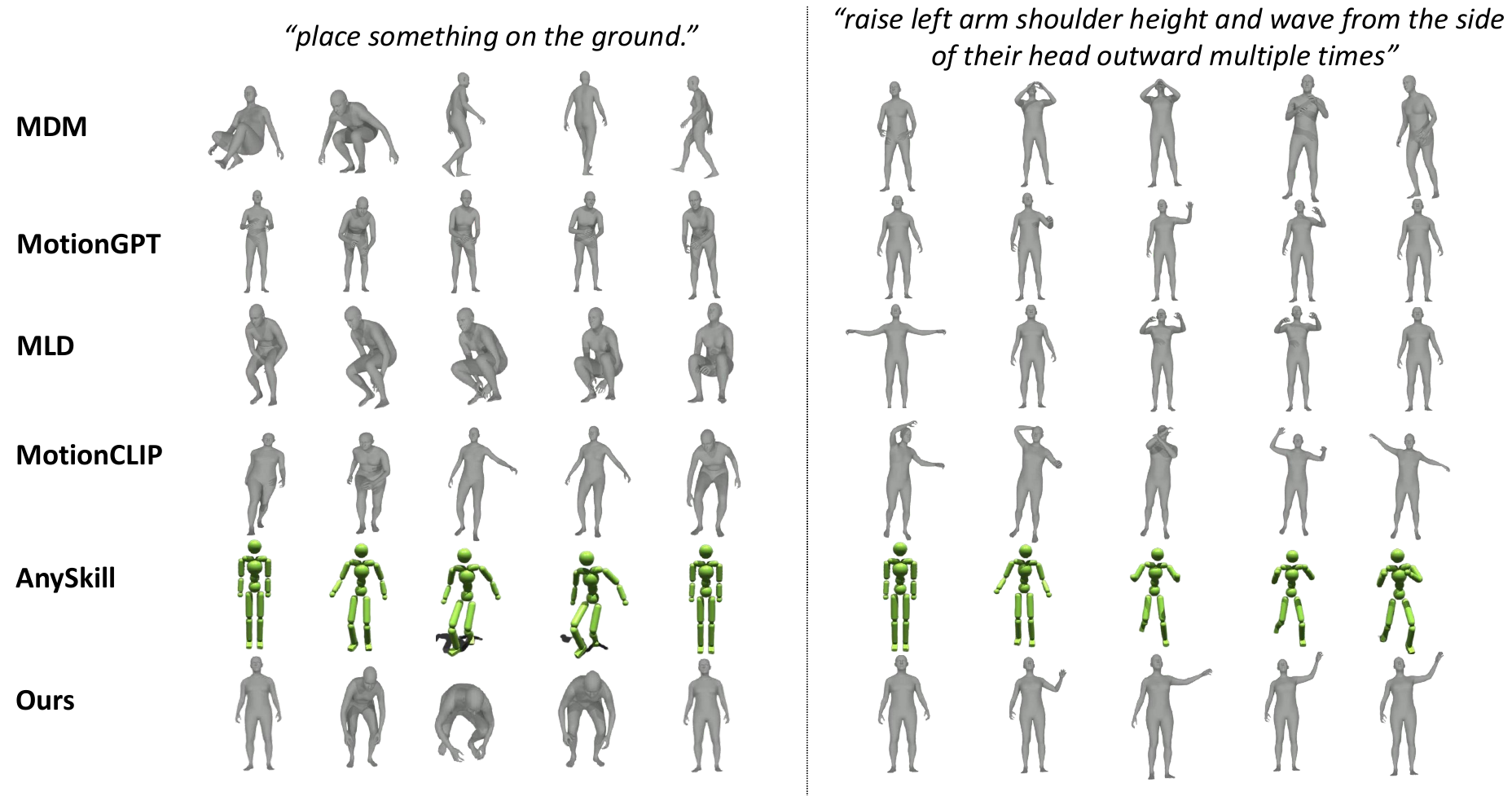}
    \caption{Additional qualitative comparison results of motions generated by different methods.}
    \label{fig:add_res_2}
\end{figure*}

\begin{figure*}[t]
    \centering
    \includegraphics[width=0.99 \linewidth]{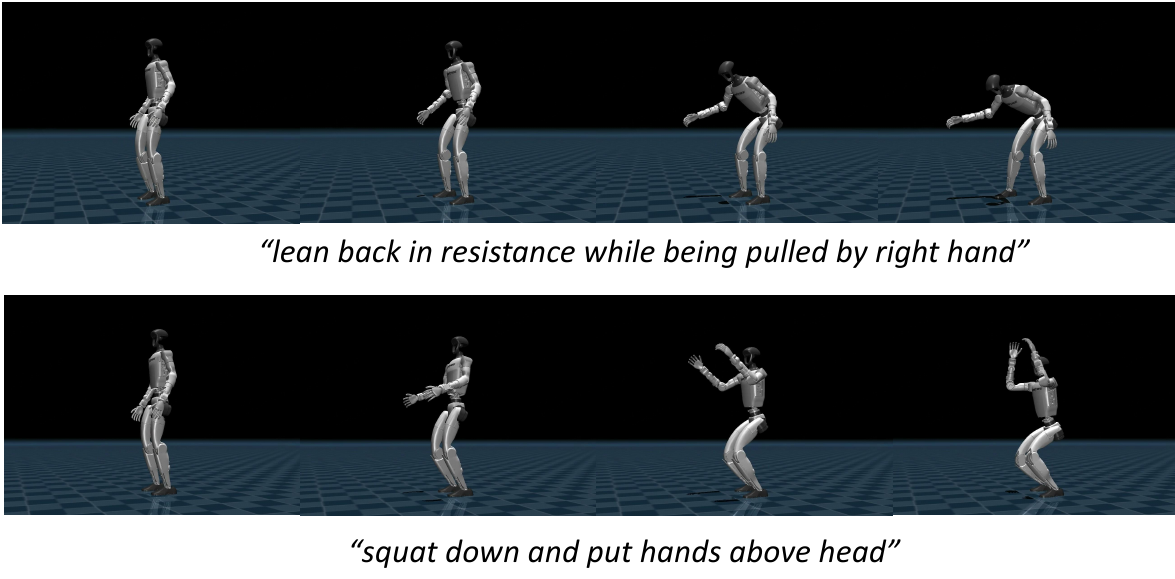}
    \caption{Additional results on the MuJoCo platform.}
    \label{fig:add_mujoco}
\end{figure*}

\subsection{Additional Visualization Results}

We present additional qualitative results comparing the motions generated by our method with those of other baselines in Fig.~\ref{fig:add_res_1} and Fig.~\ref{fig:add_res_2}. It can be observed that our approach achieves superior performance. For instance, given the description ``place something on the ground'', MDM generates motions with meaningless walking. MotionGPT shows the intention of placing an object, but fails to put it on the ground. MLD, MotionCLIP and AnySkill produce ambiguous motions that don't convey the description. In contrast, our method accurately demonstrates the complete process of placing an object on the ground and standing back up as described.

Furthermore, we present additional results of the G1 robot on the MuJoCo platform in Fig.~\ref {fig:add_mujoco}, imitating motions generated by our method, which further demonstrates the practical applicability of our approach.
\end{document}